\documentclass[letterpaper,times]{IONconf-v2}
\usepackage[hidelinks]{hyperref}

\usepackage{float}
\usepackage{mathtools}
\usepackage{amssymb}
\usepackage{bm}
\usepackage{bbm}
\usepackage{algorithm}
\usepackage{algpseudocode}
\usepackage[table,xcdraw]{xcolor}
\usepackage{tabto}
\usepackage{nomencl}
\makenomenclature

\title{Greedy Detection and Exclusion of Multiple Faults using Euclidean Distance Matrices}

\author[1]{Derek Knowles}

\author[2]{Grace Gao*}

\address[1]{Department of Mechanical Engineering, Stanford University}

\address[2]{Department of Aeronautics and Astronautics, Stanford University}

\corres{*Grace Gao\\ 
Department of Aeronautics and Astronautics, Stanford University\\
Stanford, CA, USA, 94305\\
\email{gracegao@stanford.edu}}

\begin{document}
\newcommand{\todo}[1]{{\color{red}TODO: #1}}
\newcommand{\reva}[1]{{\color{red} #1}}

\newcommand{\gnsslibpy}{\texttt{gnss\_lib\_py}}

\newcommand{\D}{\textbf{D}}
\newcommand{\X}{\textbf{X}}
\newcommand{\Gr}{\textbf{G}}
\newcommand{\J}{\textbf{J}}
\newcommand{\Dm}{\textbf{D}}
\newcommand{\Grm}{\textbf{G}}

\newcommand{\inv}{^{-1}}

\abstract[Abstract]{
Numerous methods have been proposed for global navigation satellite system (GNSS) receivers to detect faulty GNSS signals. 
One such fault detection and exclusion (FDE) method is based on the mathematical concept of Euclidean distance matrices (EDMs).
This paper outlines a greedy approach that uses an improved Euclidean distance matrix-based fault detection and exclusion algorithm.
The novel greedy EDM FDE method implements a new fault detection test statistic and fault exclusion strategy that drastically simplifies the complexity of the algorithm over previous work.
To validate the novel greedy EDM FDE algorithm, we created a simulated dataset using receiver locations from around the globe. The simulated dataset allows us to verify our results on 2,601 different satellite geometries.
Additionally, we tested the greedy EDM FDE algorithm using a real-world dataset from seven different android phones.
Across both the simulated and real-world datasets, the Python implementation of the greedy EDM FDE algorithm is shown to be computed an order of magnitude more rapidly than a comparable greedy residual FDE method while obtaining similar fault exclusion accuracy.
We provide discussion on the comparative time complexities of greedy EDM FDE, greedy residual FDE, and solution separation.
We also explain potential modifications to greedy residual FDE that can be added to alter performance characteristics.
}

\maketitle

\section{Introduction}

The number of satellites available for positioning, navigation, and timing (PNT) has continued to increase over the years. As an example, we used the open-source library, \gnsslibpy{} \citep{knowles2023gnss}, to create Figure \ref{fig:denver_skyplot} which shows that over an afternoon in downtown Denver, Colorado on August, 1$^{\text{st}}$ 2023, measurements from at least 48 PNT satellites were available in open-sky conditions. The advent of LEO constellations will only increase the number of available PNT satellites \citep{kassas2023navigation}. Due to the increasing number of satellites, it is imperative that fault detection and exclusion methods are computed rapidly in the presence of many satellite measurements and multiple faults. In this paper, measurement faults encompass any pseudorange biases caused by factors such as multipath, non-line-of-sight signals, atmospheric effects, satellite faults, etc.

Many fault detection and exclusion (FDE) methods have been proposed in literature over the years. Two of the most common methods are solution separation and residual methods \citep{PARKINSON1988, Pullen2021, Ma2019, Blanch2012, Joerger2014, Zhu2018a}. 
Solution separation methods exclude faulty measurements by computing a position solution with all satellite measurements and comparing that position solution with other position solutions computed by using a subset of all the satellite measurements available. However, Figure \ref{fig:combos} shows that creating combinations of measurements scales in a combinatorial manner as both the number of measurements and the number of faulty measurements increases. If we have 48 measurements in Denver and want to remove eight faults, then we would need to compute $3.7 \times 10^8$ different subset combinations, and if we only want to use the 20 satellites that are the most self-consistent among each other, then there are $1.6 \times 10^{13}$, or 16~trillion, different subset combinations that we would need to form and test if using solution separation FDE. Solution separation is an excellent algorithm choice with some optimality guarantees \citep{blanch2015fast} for cases similar to aviation where the open-sky condition of the receiver affords few faults and pseudorange measurements closely follow a known error distribution \citep{blanch2015,joerger2016}. 
However, the combinatorial search of solution separation is computationally intractable in conditions where there are many measurements and many faults and also performs more poorly in our tested scenarios with 10s to 100s of meters of pseudorange residuals.
For these reasons, we provide a theoretical runtime complexity comparison against solution separation in Section~\ref{app:time} but do not compare against solution separation in the experimental results.
The computational intractability of solution separation and other similar combinatorial search FDE methods has given rise to the search for faster methods.

Several methods have been proposed to increase the speed of FDE by eliminating the combinatorial search for a consistent set of measurements. The greedy residual method computes the pseudorange residuals using all satellite measurements and then removes the single satellite with the largest normalized residual until the chi square test statistic falls below the provided threshold \citep{blanch2015fast}.
This algorithm is in the class of algorithms called greedy since it tries to find a globally optimal solution by making a locally optimal choice in each iteration \citep{greedy_algorithms}.
This greedy residual method is much faster than solution separation, but a new position estimate must be computed for each greedy iteration which can be computationally expensive. Another method creates a non-iterative approach by calculating residuals at each point in a four-dimensional grid representing the solution space of the receiver's position and clock bias rather than in the measurement domain \citep{wendel2022gnss}.
Rather than scaling with the number of faults, this method scales with increasing granularity of the receiver's solution space grid. One downside of the algorithm is that the solution space grid is assumed to include the receiver's true position. This assumption means that a user must already have a nearly accurate position estimate at which the solution space grid is centered. Other methods have looked at alternate options of greedily choosing measurement subsets \citep{zhang_fast}.

\begin{flushleft}     
\begin{minipage}[t]{0.477\textwidth}

\begin{figure}[H]
    \raggedleft
    \centering
    \includegraphics[width=\linewidth]{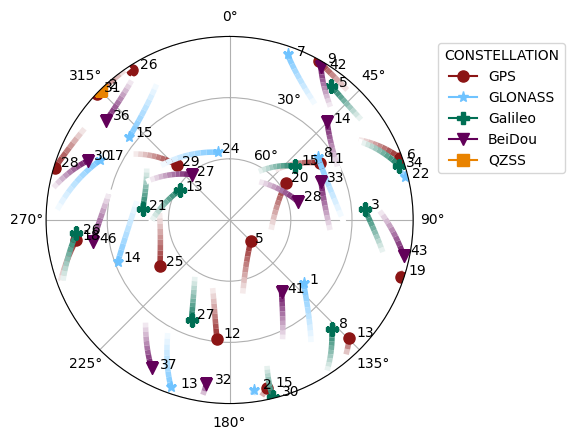}
    \captionof{figure}{Over 48 positioning, navigation, and timing satellites are in view over Denver, Colorado.}
    \label{fig:denver_skyplot}
\end{figure}

\end{minipage}%
\hfill
\begin{minipage}[t]{0.503\textwidth}

\begin{figure}[H]
    \raggedleft
    \centering
    \includegraphics[width=\linewidth]{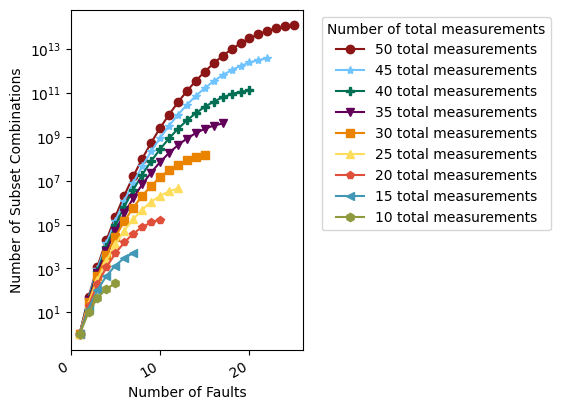}
    \captionof{figure}{Number of subset combinations required by solution separation FDE depending on the number of measurements and faults.}
    \label{fig:combos}
\end{figure}

\end{minipage}%
\end{flushleft}

Another FDE approach removes the combinatorial computation time of fault detection and exclusion by using a mathematical tool called Euclidean distance matrices (EDMs).
This EDM FDE method has previously been shown to run using less compute time while maintaining high accuracy in both detecting and excluding faults \citep{knowles_navigation}. Additionally, EDM FDE does not explicitly require an estimate of the receiver's position to be able to perform FDE.

In the last two years since EDM  FDE was originally presented, we have extensively tested the algorithm across multiple GNSS datasets.
Additionally, we have gained additional insight into the mathematical basis for the algorithm by further inspecting the impact on the rank properties of EDMs when an EDM is constructed from faulty versus non-faulty signals.
In this paper, we present a novel EDM FDE method that rapidly removes multiple faults by greedily removing the single satellite with the largest bias/fault in each iteration.
Similar to greedy residual FDE, our EDM algorithm is also called greedy since it takes a locally optimal decision to remove the satellite with the largest bias/fault in each iteration in hopes of finding the globally optimal solution~\citep{greedy_algorithms}.
We rigorously validate the advantages of the greedy EDM FDE method over a comparable greedy residual FDE method.
This paper is based on our ION GNSS+ conference paper \citep{knowles2023detection} but adds experimental results on a real-world dataset, additional insight on the impact of the receiver clock bias, and a detailed time complexity comparison against residual-based FDE and solution separation FDE in Section~\ref{app:time}.

We first present a brief background on Euclidean distance matrices used for GNSS fault detection and exclusion in Section \ref{sec:background}.
Next, we explain the details of the new greedy EDM fault detection and fault exclusion strategies in Section \ref{sec:approach} and explicitly describe its main differences over previously proposed EDM FDE methods \citep{knowles_navigation}.
In Section \ref{sec:simulated}, we describe a simulated dataset we created to compare our greedy EDM FDE method with greedy residual FDE in terms of computation time and fault exclusion accuracy.
In Section~\ref{sec:gsdc}, we compare greedy residual FDE and our previous version of EDM FDE \citep{knowles_navigation,githuboldedm} on a real-world dataset. Using the real-world dataset, we present results on computation time and localization accuracy for each method after excluding faulty measurements.
Both greedy EDM FDE and greedy residual FDE are implemented within the open-source GNSS Python library \gnsslibpy{}~\citep{knowles2023gnss,knowles2022modular} which allows users to easily test the FDE methods on their own GNSS data.
Section \ref{app:time} provides a detailed theoretical runtime comparison between greedy EDM FDE, greedy residual FDE, and solution separation FDE.
Finally, Section \ref{sec:mods} describes several possible modifications to greedy EDM FDE to change performance characteristics.
The source code for all experiments and figures are available open source \citep{githubgreedy}.

\section{Euclidean Distance Matrix Background}
\label{sec:background}

\begin{figure}[H]
    \begin{subfigure}[]{0.4\textwidth}
        \label{fig:gnss_tetra}
        \centering 
        \includegraphics[height=5cm]{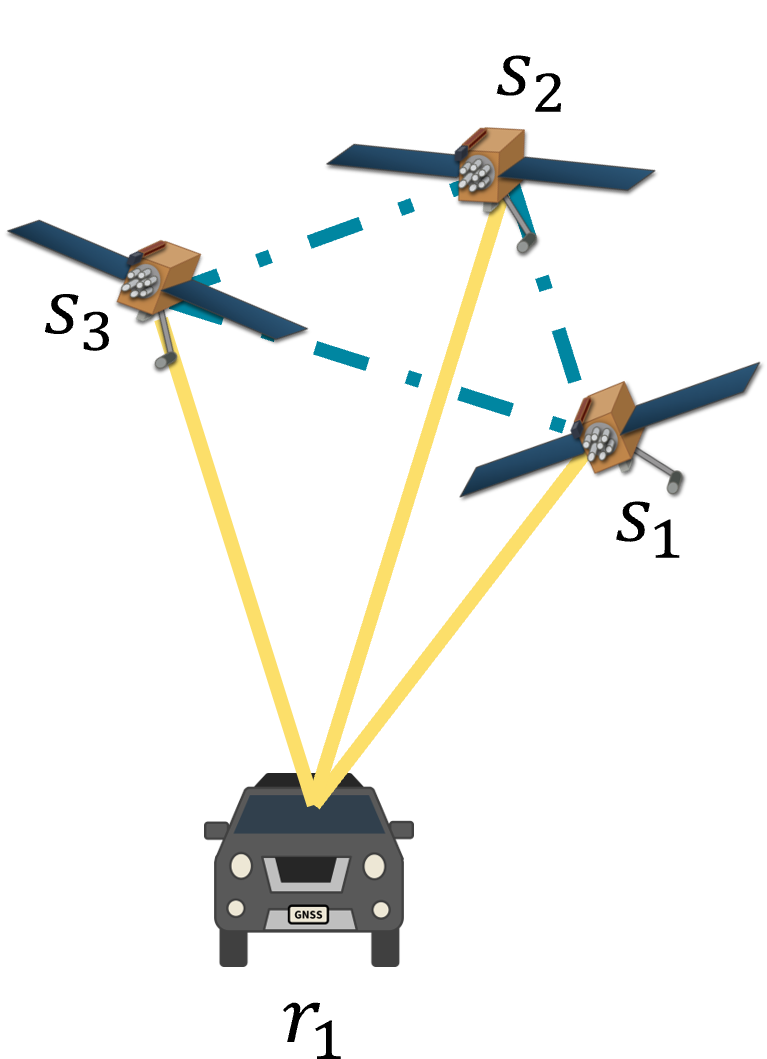} 
        \caption{}
    \end{subfigure}
	\hspace{0.2\textwidth}
    \begin{subfigure}[]{0.2\textwidth}
	  \label{fig:gnss_sources}
        \centering 
        \includegraphics[height=5cm]{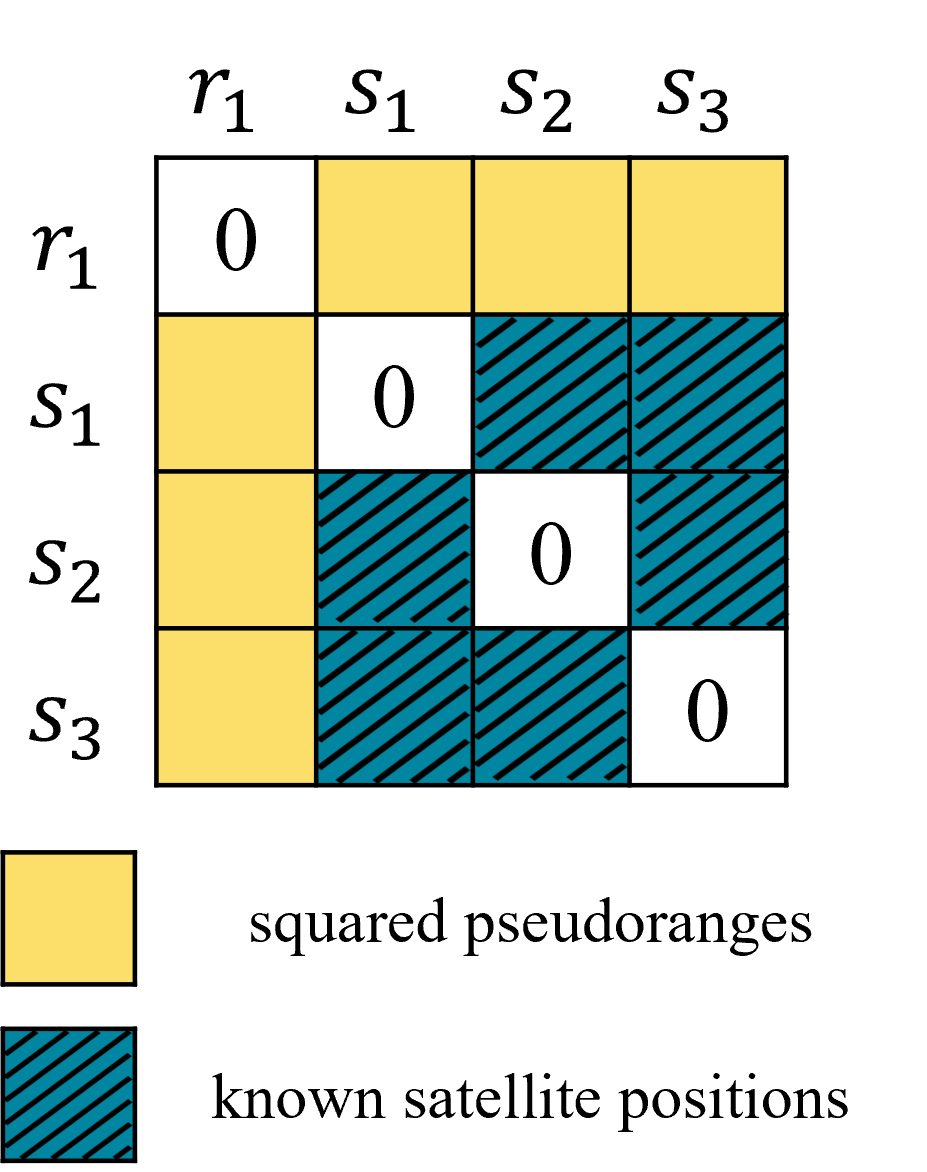}
        \caption{}
        \end{subfigure}
	\caption{A combination of squared pseudoranges and known satellite positions are used to construct the EDM shown in subfigure (b) for the system shown in subfigure (a) of a single receiver and three satellites. Subfigure (b) taken from \cite{knowles_navigation}.}
	\label{fig:gnss_formulation}
\end{figure}

Euclidean distance matrices are a mathematical tool used for signal processing and obtaining information from distances~\citep{Dokmanic2015}. If constructed in a specific manner, Euclidean distance matrices are also able to check for GNSS signal faults thanks to a special rank property the matrices possess. By definition, Euclidean distance matrices contain the squared distances between all sets of points in a system. Figure \ref{fig:gnss_formulation} shows a visualization of how a Euclidean distance matrix is formed for a simple system comprised of three satellites and a single receiver. The distances between satellites are calculated using satellite ephemeris and the distances between the satellites and receiver are provided through the pseudorange measurement. A more complete background of Euclidean distance matrices and their relevance for GNSS fault detection is found in \cite{knowles_navigation}.
Once we construct the EDM, \Dm, using processed pseudoranges and known satellite positions, we retrieve its corresponding symmetric Gram matrix, \Grm, by double centering the EDM through Equation (\ref{eq:center}) where $\textbf{I}$ represents the identity matrix, \textbf{1} is a column vector of ones, and $n$ is three --- the dimension of the space spanned by the points in the system. For more information on recovering a Gram matrix through double centering, refer to (\cite{Dattorro}, section 5.4.2.2).
\begin{equation}
    \label{eq:center}
    \text{\Grm} = -\frac{1}{2} \J \text{\Dm} \J \text{\quad where } \J = \textbf{I} - \frac{1}{n} \textbf{1} \textbf{1}^\top.
\end{equation}
 If all distances are perfectly consistent, then the Gram matrix, \Grm, has rank equal to the dimension of the state space \citep{Horn1985}. In the case where we are solving for a three-dimensional receiver position, the state space has dimension of three, i.e. $n = 3$. We explain in Section \ref{sec:approach} how the rank of the Gram matrix, \Grm, changes when there are faulty measurements inconsistent with the rest of the measurements.
 
\section{Approach}
\label{sec:approach}

In this section, we explain our approach for using EDMs to perform fault detection and exclusion. Our approach uses the particular geometry of the GNSS positioning problem to apply Euclidean distance matrices in a novel manner. This approach uses a different fault detection test statistic and fault exclusion strategy than previous work in \citep{knowles_navigation}. The FDE approach has been drastically simplified thanks to increased understanding of EDMs and their use for fault detection and exclusion. In Section~\ref{sec:detection}, we explain our EDM fault \textit{detection} approach and in Section~\ref{sec:exclusion}, we explain our greedy EDM fault \textit{exclusion} approach. In Section~\ref{sec:edm_improvements} we explicitly describe the improvements of this work over previous EDM FDE methods, and in Section~\ref{sec:reproduce} we illustrate how all experiments can be reproduced.

\subsection{Fault Detection}
\label{sec:detection}

The key detail for using the recovered Gram matrix for fault detection is understanding how the eigenvalues of the Gram matrix change when faults are present in the measured pseudoranges used to construct the GNSS EDM. With increasing fault magnitude, the following happens to the eigenvalues as ordered from largest to smallest:
\begin{enumerate}
    \item The first $n$ eigenvalues remain a consistent value. See Section~\ref{sec:background} for the definition of $n$.
    \item The $n+1$ and $n+2$ eigenvalues increase \textit{together as a pair}.
    \item The $n+3$ and all subsequent eigenvalues remain a consistent value.
\end{enumerate}

The eigenvalues of the Gram matrix are an indication of its rank. When the $n+1$ and subsequent eigenvalues are nearly zero, that is an indication that the Gram matrix is close to having a rank of $n$. In Section \ref{sec:background}, we clarified that all valid Gram matrices should have rank of $n$. When a fault is present in the pseudorange measurements and the $n+1$ and $n+2$ eigenvalues increase together as a pair, that means that the Gram matrix has rank of $n+2$ instead of rank $n$ as we expected.

One main insight and difference over previous work is to use \textit{the pair} of the $n+1$ and $n+2$ eigenvalues for the fault detection test statistic. Since neither the first $n$ nor the $n+3$ and subsequent eigenvalues change significantly with increasing fault magnitude, none of those eigenvalues should be used for fault detection. The new fault detection test statistic is shown in Equation \ref{eq:detect}.
\begin{equation}
    \label{eq:detect}
    \frac{\lambda_{n+1} + \lambda_{n+2}}{2\lambda_1} > \text{detection threshold}
\end{equation}
The EDM fault detection test statistic averages the $n+1$ and $n+2$ eigenvalues and divides by the largest eigenvalue to normalize the test statistic between zero and one and make the test statistic relatively agnostic to the amount of natural noise in the pseudorange measurements. If the test statistic is larger than the provided threshold then the measurements are deemed inconsistent and a fault is predicted within the measurements. However, if the test statistic is smaller than the provided threshold, then the measurements are deemed consistent and no faults are predicted within the measurements. The threshold is set by the user based on the natural noise present in the receiver's pseudorange measurements. With increasing natural noise in the pseudorange measurements, the detection threshold should also increase so as to not remove non-faulty measurements.

\begin{figure}[H]
    \centering
    \includegraphics[width=0.7\linewidth]{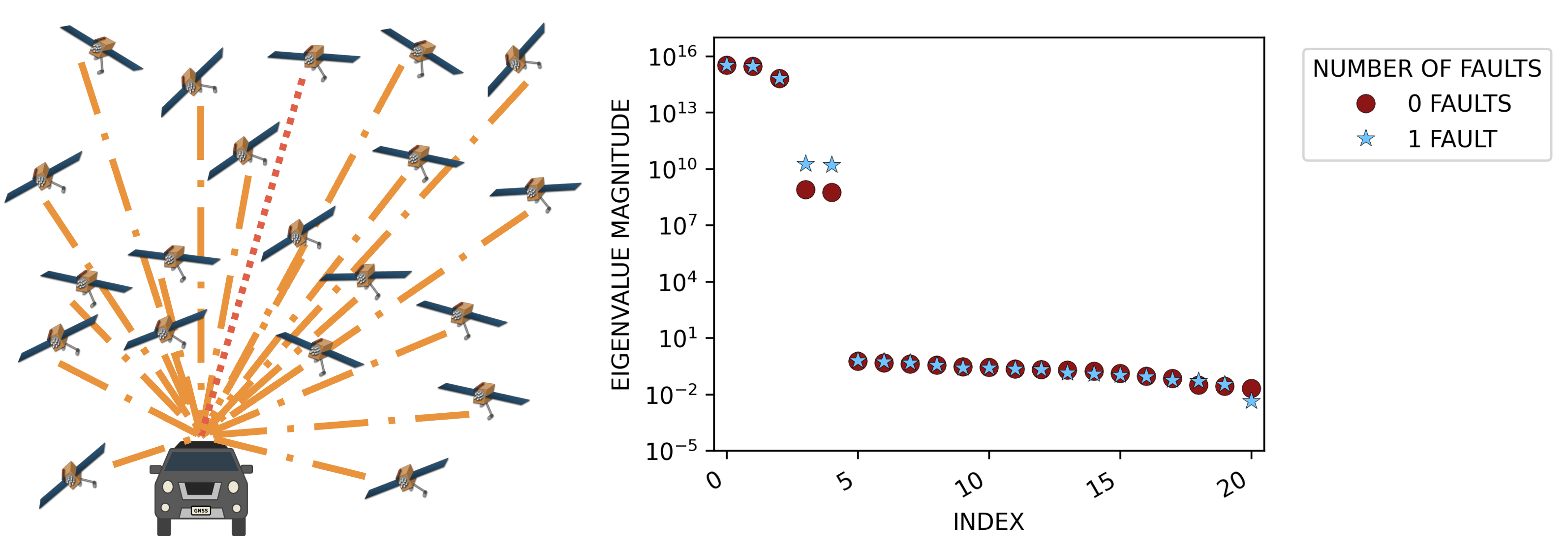}
    \captionof{figure}{Even with noise, a single faulty bias added to a measurement dramatically increases the fourth and fifth eigenvalue of the Gram matrix.}
    \label{fig:detection}
\end{figure}

The proposed fault detection test statistic works even when zero mean Gaussian noise is present in the pseudorange measurements. Figure~\ref{fig:detection} shows an example of a timestep with $n=3$ and 20 received satellite measurements. With zero mean Gaussian noise added to all the pseudoranges, the fourth and fifth eigenvalues are nonzero even when there are no faults present in the pseudoranges. However, when even a single fault exists in the measurements, the fourth and fifth eigenvalues still drastically jump an order of magnitude. This means that even in the presence of noisy pseudorange measurements, we can still accurately detect the existence of faulty signals by detecting a jump in the value of the fourth and fifth eigenvalues.
\subsection{Fault Exclusion}
\label{sec:exclusion}

For fault exclusion, we inspect the eigenvectors associated with both the $n+1$ and $n+2$ eigenvalues. The intuition is that our fault exclusion strategy identifies the satellite measurements that are most influencing the increasing magnitude of the $n+1$ and $n+2$ eigenvalues. In implementation, we take the absolute value of the $n+1$ and $n+2$ eigenvectors and then take the elementwise average across the two vectors. The measurement with the largest average absolute value is predicted to be a measurement fault. Figure \ref{fig:q} shows an example of the absolute value of the eigenvectors, $\textbf{Q}$, obtained through eigenvalue decomposition of the Gram matrix, \Grm, with $n=3$. In the example measurements, there exists a single measurement fault from the fourth satellite, $s_4$. The measurements from the fourth satellite produce the largest absolute values in the fourth and fifth eigenvectors indicating that it is the cause of the detected fault. Note that the first row of the $\textbf{Q}$ matrix shown in Figure~\ref{fig:q} corresponds to the receiver's index since the EDM was constructed according to the order shown in Figure~\ref{fig:gnss_formulation}.
The first row also has a relatively large absolute value magnitude indicating the fact that the receiver's position is directly influenced by the faulty satellite pseudorange measurement.

Similar to greedy residual FDE's approach of greedily removing the measurement with the largest normalized residual in each iteration \citep{blanch2015fast}, our greedy EDM FDE algorithm greedily chooses the satellite measurement to remove at each iteration with the largest effect on the $n+1$ and $n+2$ eigenvectors. This greedy removal of satellite measurements happens iteratively until the fault detection test statistic falls below the provided threshold. Algorithm \ref{alg:edm-fde} shows the full greedy EDM FDE algorithm.

\algnewcommand{\Inputs}[1]{%
  \State \textbf{Inputs:}
}
\algnewcommand{\Initialize}[1]{%
  \State \textbf{Initialize:}
}

\algdef{SE}[SUBALG]{Indent}{EndIndent}{}{\algorithmicend\ }%
\algtext*{Indent}
\algtext*{EndIndent}

\begin{flushleft}     
\begin{minipage}[b]{0.55\textwidth}

\begin{figure}[H]
    \raggedleft
    \centering
    \includegraphics[height=6.5cm]{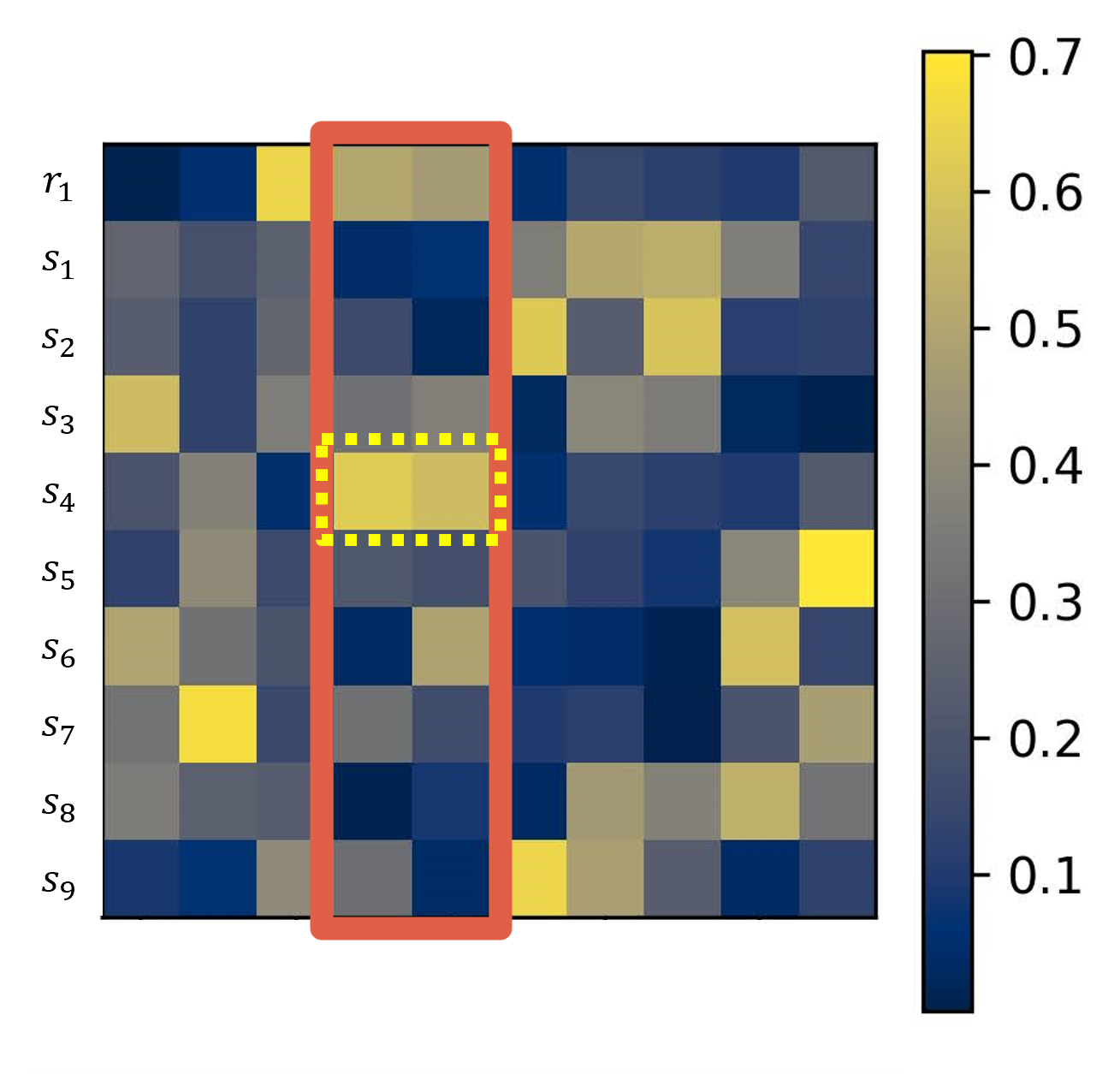}
    \caption{Eigenvectors, $\textbf{Q}$, of the Gram matrix, \Grm.
    }
    \label{fig:q}
\end{figure}

\end{minipage}%
\hfill
\begin{minipage}[b]{0.4\textwidth}
	\begin{algorithm}[H]
    	\caption{Greedy EDM FDE}
    	\label{alg:edm-fde}
    	\begin{algorithmic}[1]
    	    
            \Inputs{}
                \Indent
                \State $n$ : Dimension of state space
                \State \Dm : Euclidean distance matrix
                \State $T_d$ : detection threshold
                \EndIndent
            \Initialize{}
                \Indent
                \State \text{\Grm}$ \leftarrow -\frac{1}{2}\textbf{J}\text{\Dm}\textbf{J}$
                \State $\textbf{Q},\boldsymbol\Lambda \leftarrow \text{EVD}(\text{\Grm})$
                \EndIndent
    		\While{\# satellites $>$ 4}
                \If{$(\lambda_{n+1} + \lambda_{n+2})/(2\lambda_1) < T_d$}
                \State \textbf{break}
                \EndIf
    		    \State index $\leftarrow \max\left(\left|(q_{n+1} + q_{n+2})/2\right|\right)$
    		    \State exclude index from \Dm
                \State $\text{\Grm} \leftarrow -\frac{1}{2}\textbf{J}\text{\Dm}\textbf{J}$
                \State $\textbf{Q},\boldsymbol\Lambda \leftarrow \text{EVD}(\text{\Grm})$
    		\EndWhile
    	\end{algorithmic}
    \end{algorithm}
\end{minipage}%
\end{flushleft}

\subsection{Improvements over Previous EDM FDE Methods}
\label{sec:edm_improvements}
The approach described in Section~\ref{sec:detection} and Section~\ref{sec:exclusion} is a significant improvement over our prior work in \cite{knowles_navigation}.
Throughout the rest of the paper, we refer to our previous EDM FDE algorithm as our EDM 2021 algorithm since that was the year in which the method was originally published \citep{githuboldedm}.
Our EDM 2021 algorithm used Equation~\ref{eq:detect_old} for the fault detection test statistic where $m$ is the total number of measurements available. 
\begin{align}
    \label{eq:detect_old}
    \frac{\lambda_{n+1} \times \frac{1}{m-n} \sum^m_{i = n+1} \lambda_i}{\lambda_1} > \text{detection threshold}
\end{align}
Based on new experience working with EDM matrices, we discovered the relationship shown in Figure~\ref{fig:detection} that the $n+1$ and $n+2$ eigenvalues increase together as a pair when a fault is present. While Equation~\ref{eq:detect_old} from previous work multiplied the $n+1$ and the average of the $n+1$ to $m$ eigenvalues, we now know that the $n+3$ and subsequent eigenvalues remain a consistent value so their magnitudes do not add significance to test statistic. Hence, in our updated test statistic shown in Equation~\ref{eq:detect}, we only look at the average of the $n+1$ and $n+2$ eigenvalues which affords a more clear signal to recognize faults with the test statistic.

Another significant improvement over prior work is the simplicity of the new fault exclusion approach. The simple method from Algorithm \ref{alg:edm-fde} can be implemented directly with accurate detection performance. The EDM 2021 implementation of the algorithm used a complex check iterating through possible candidates of both the maximums and minimums of the left and right singular vectors of the Gram matrix \citep{githuboldedm}. Based on our new experience, we now know to inspect the absolute value of the $n+1$ and $n+2$ eigenvectors which dramatically simplifies the implementation of Algorithm~\ref{alg:edm-fde} which is available open source~\citep{githubgreedy}.

Finally, in this work we test against both simulated and a real-world dataset with the algorithm fully implemented in an open-source library. All of our results are reproducible as described in Section~\ref{sec:reproduce}.

\subsection{Algorithm Reproducibility}
\label{sec:reproduce}
\begin{figure}[H]
    \centering
    \includegraphics[height=7cm]{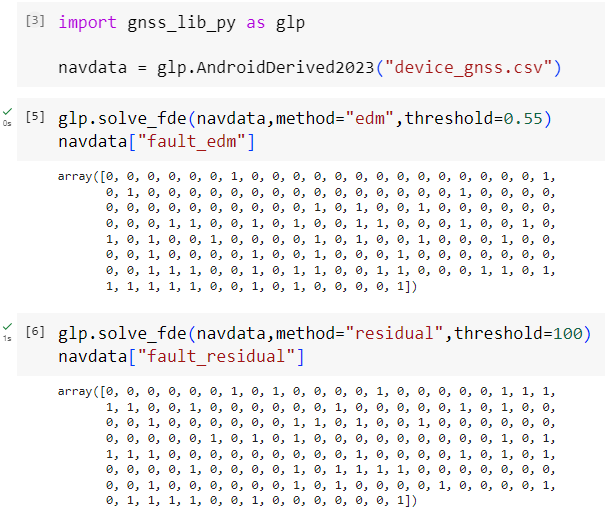}
    \captionof{figure}{Example function calls to run either EDM FDE or residual FDE.}
    \label{fig:edm_code}
\end{figure}

To enable the reproducibility of all experiments, both greedy EDM FDE and the baseline comparison of greedy residual FDE have been implemented in our open-source GNSS Python library \gnsslibpy{} \citep{knowles2023gnss,knowles2022modular}. Figure~\ref{fig:edm_code} shows that with just a few lines of code, users can upload data from the Google Smartphone Decimeter Challenge 2023 \citep{smartphone-decimeter-2023} and then perform either EDM or residual FDE simply by changing the method input parameter to the function call. The output of the function returns a one if a fault is predicted and a zero if no fault is predicted for each satellite measurement in the dataset.

For more detailed instructions, tutorials are available on how to use the \gnsslibpy{} library and its associated FDE functions at \href{https://gnss-lib-py.readthedocs.io/}{gnss-lib-py.readthedocs.io}. All specific code to run the experiments in Section~\ref{sec:simulated} and Section~\ref{sec:gsdc} as well as generate the included figures is also open source \citep{githubgreedy}.

\section{Simulated Dataset Results}
\label{sec:simulated}
To initially validate our greedy EDM FDE approach, we created a simulated dataset using nine global locations: Calgary, Canada; Cape Town, South Africa; Hong Kong; London, United Kingdom; Munich, Germany; Sao Paulo, Brazil; San Francisco, USA; Sydney, Australia; and Zurich, Switzerland. At each location, we simulated measurements at five minute intervals across a time span of 24~hours. We used the open-source library \gnsslibpy{} \citep{knowles2023gnss} to obtain satellite locations given a specific timestamp and receiver location. For three of the locations --- Calgary, London, and Zurich --- we used an elevation mask of 30~degrees to remove measurements near the horizon and for all other locations we used an elevation mask of 10~degrees.
\begin{flushleft}     
\begin{minipage}[b]{0.4925\textwidth}
\begin{figure}[H]
    \raggedleft
    \centering
    \includegraphics[width=\linewidth]{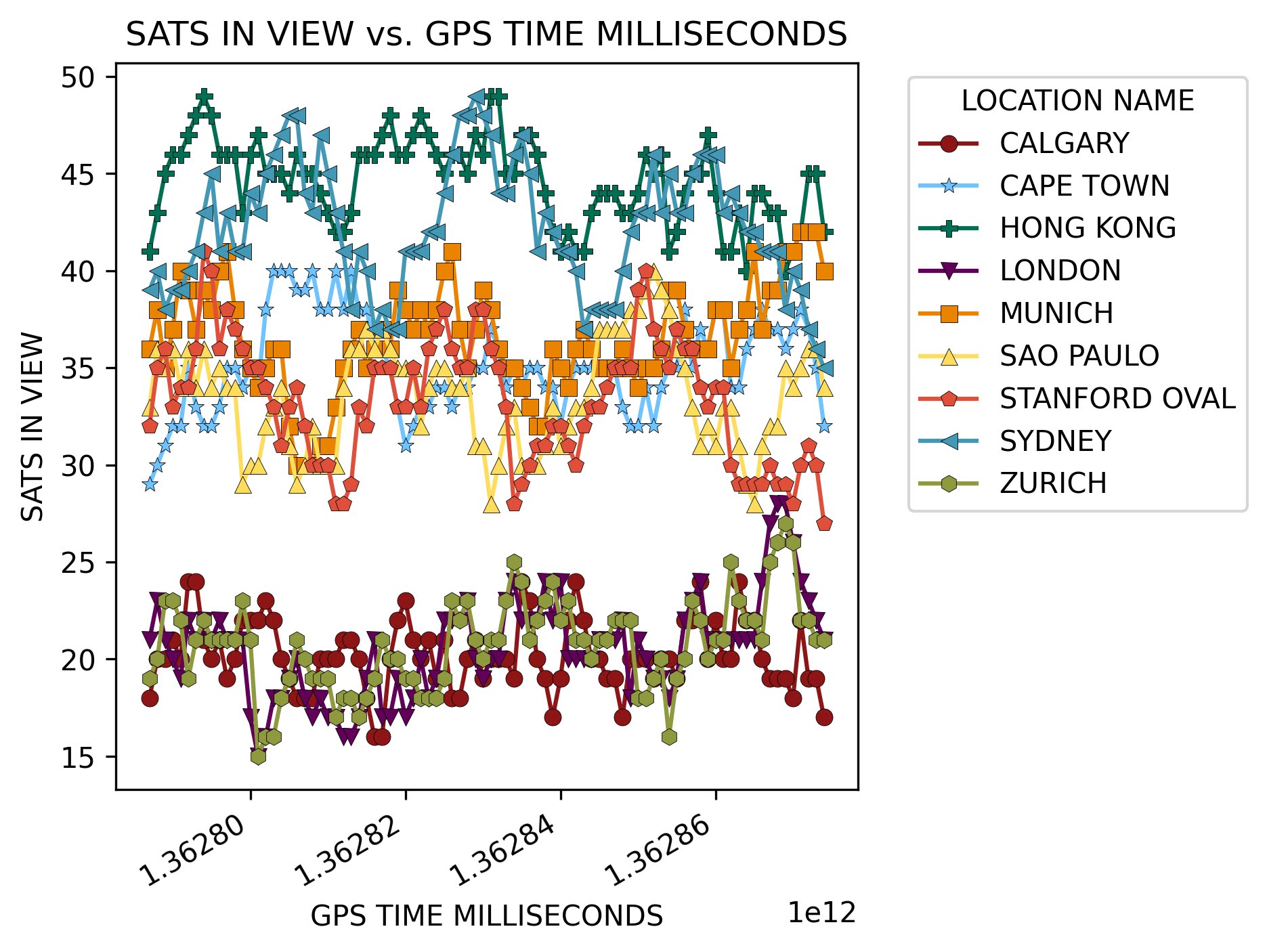}
    \captionof{figure}{Number of satellites in view across the dataset.}
    \label{fig:in_view}
\end{figure}

\end{minipage}%
\hfill
\begin{minipage}[b]{0.4875\textwidth}

\begin{figure}[H]
    \raggedleft
    \centering
    \includegraphics[width=\linewidth]{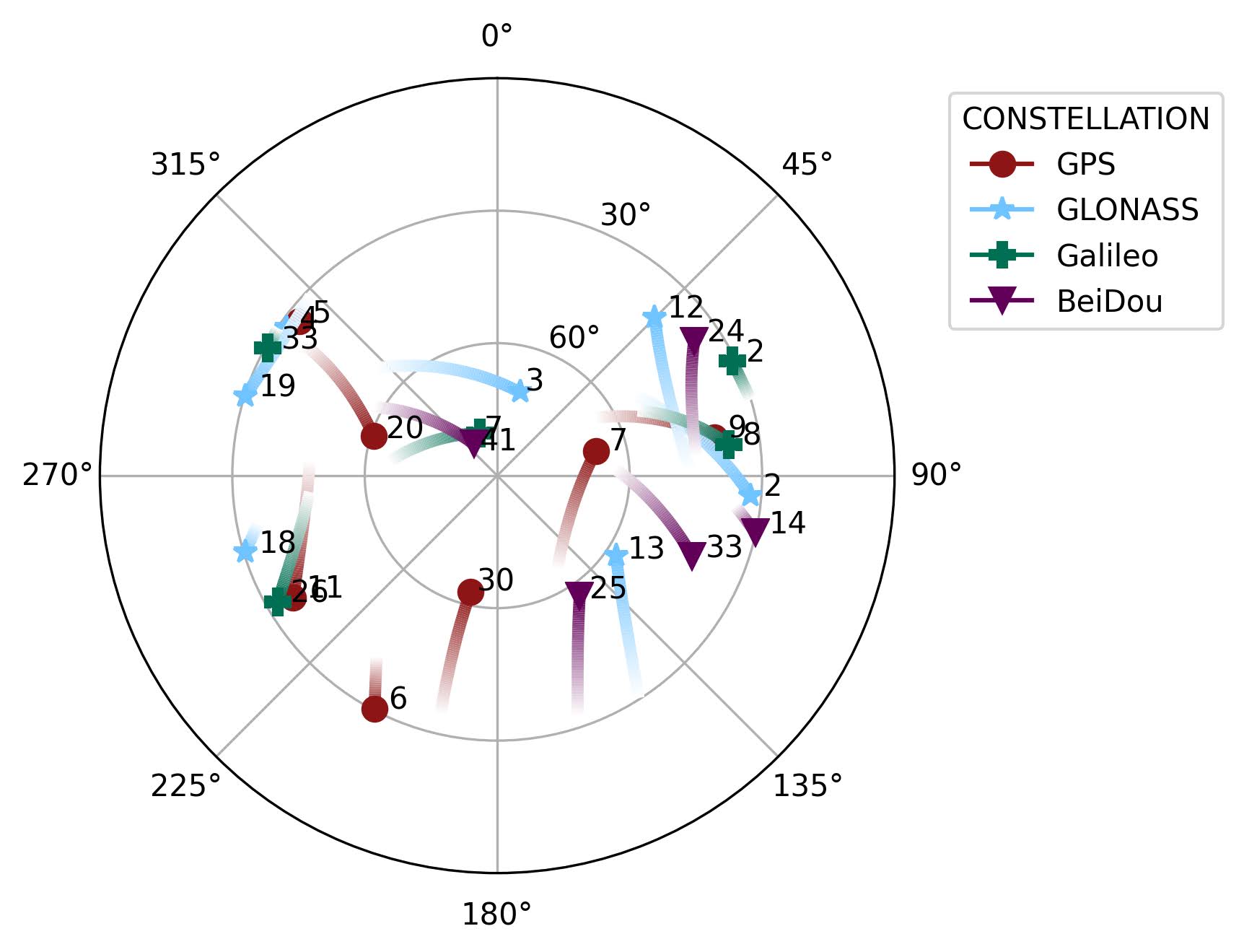}
    \captionof{figure}{Skyplot of a section of the simulated data from Zurich.}
    \label{fig:zurich_skyplot}
\end{figure}

\end{minipage}%
\end{flushleft}

\begin{flushleft}     
\begin{minipage}[t]{0.49\textwidth}

\begin{figure}[H]
    \raggedleft
    \centering
    \includegraphics[width=\linewidth]{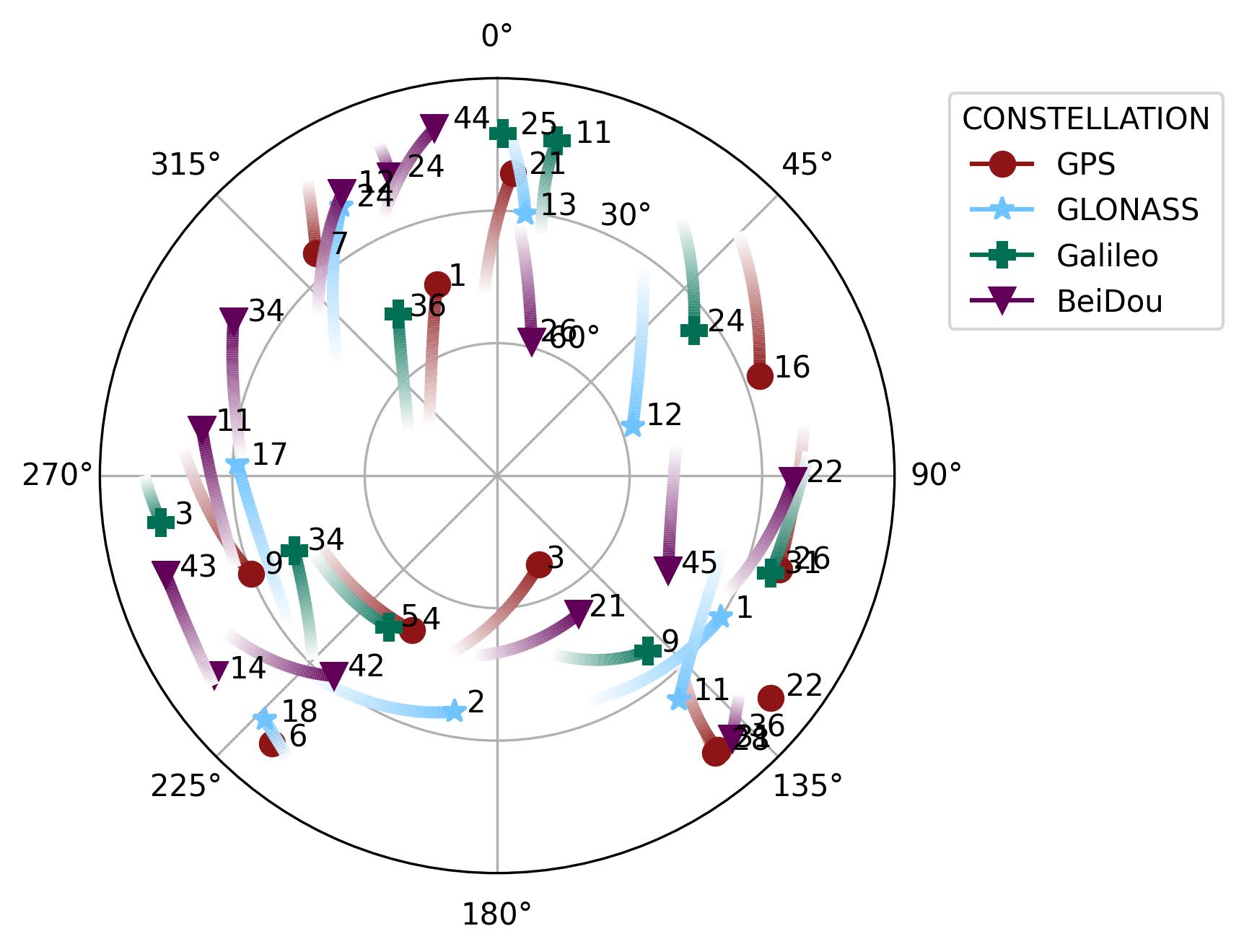}
    \captionof{figure}{Skyplot of a section of the simulated data from Sao Paulo.}
    \label{fig:sao_paulo_skyplot}
\end{figure}

\end{minipage}%
\hfill
\begin{minipage}[t]{0.49\textwidth}

\begin{figure}[H]
    \raggedleft
    \centering
    \includegraphics[width=\linewidth]{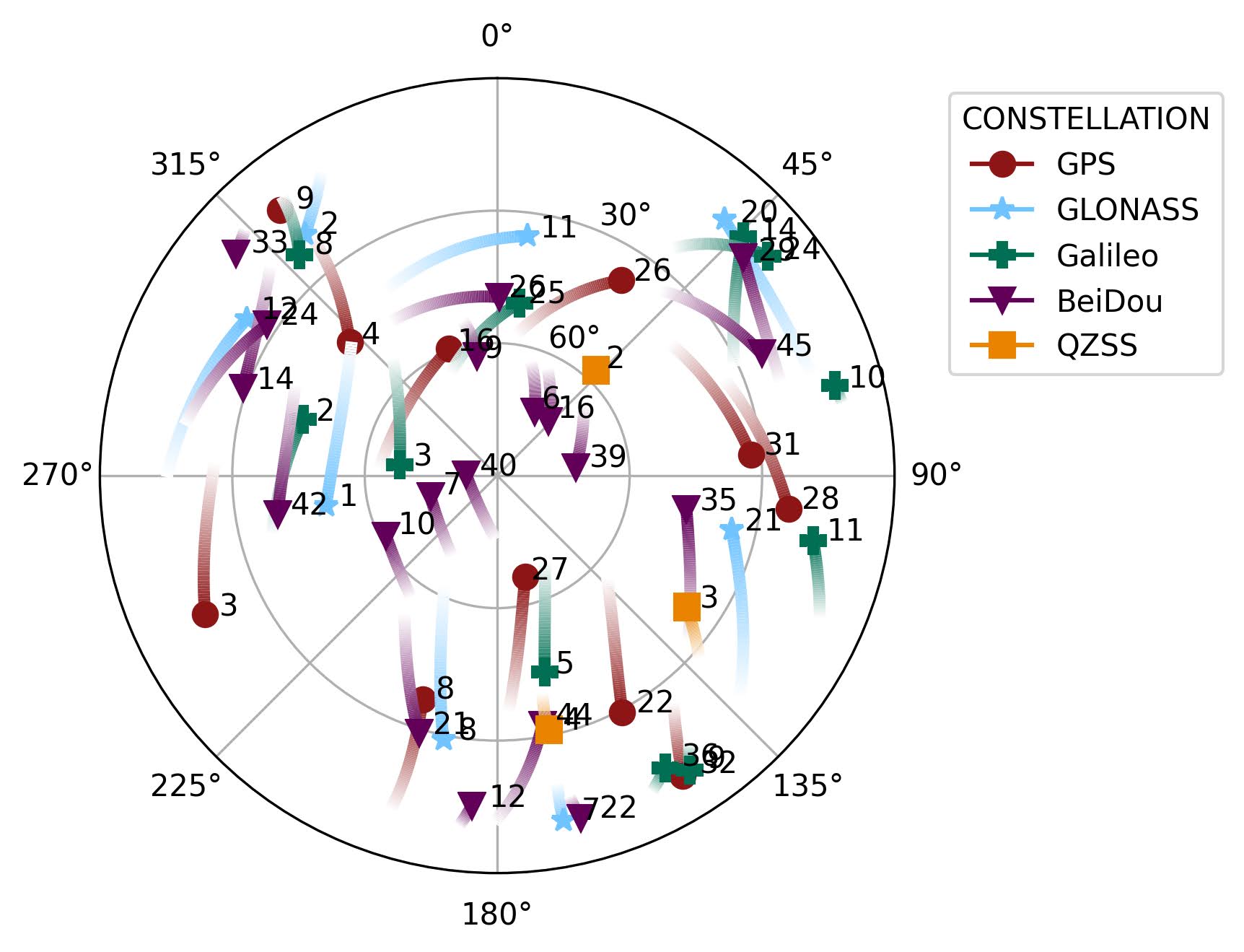}
    \captionof{figure}{Skyplot of a section of the simulated data from Hong Kong.}
    \label{fig:hong_kong_skyplot}
\end{figure}

\end{minipage}%
\end{flushleft}
Figure~\ref{fig:in_view} shows the resulting number of satellites in view at each location across the dataset. The Calgary, London, and Zurich locations have fewer satellites in view due to the larger elevation mask. In total, the simulated dataset creates 2,601 different satellite geometries to test against. Figures \ref{fig:zurich_skyplot}, \ref{fig:sao_paulo_skyplot}, and \ref{fig:hong_kong_skyplot} show the varied satellite geometries in our simulated dataset from Zurich, Sao Paulo, and Hong Kong respectively. Zero mean gaussian noise with a standard deviation of 10m was added to all measurements. Biases of 10, 20, 40, and 60 m were added to either 1, 2, 4, 8, or 12 measurements at each timestep depending on the parameters of the test.

To validate the proposed greedy EDM FDE algorithm, we compare it against greedy residual FDE on our simulated dataset.
In the Section~\ref{sec:simulated_time}, we present results of the computation time and in Section~\ref{sec:simulated_accuracy}, we present fault exclusion accuracy results.
Our results validate the rapid computation time of the greedy EDM FDE algorithm in the presence of many measurements and many faults when compared with greedy residual FDE.
\subsection{Computation Time Analysis}
\label{sec:simulated_time}
Greedy EDM FDE outperforms greedy residual FDE in terms of computation time. Figure \ref{fig:time_measurements} shows that greedy EDM FDE runs on average faster than greedy residual FDE in the presence of increasing number of measurements. Figure \ref{fig:time_faults} shows that greedy EDM FDE also runs on average faster than greedy residual FDE in the presence of increasing number of faults. In both Figure \ref{fig:time_measurements} and \ref{fig:time_faults} one standard deviation on either side of the mean computation time is shaded indicating the variability present in the computation time results. In addition to the experimental results shown in this section, Section~\ref{app:time} derives the theoretical time complexity for EDM FDE, residual FDE, and solution separation.

\begin{flushleft}     
\begin{minipage}[b]{0.46\textwidth}

\begin{figure}[H]
    \centering
    \includegraphics[width=\linewidth]{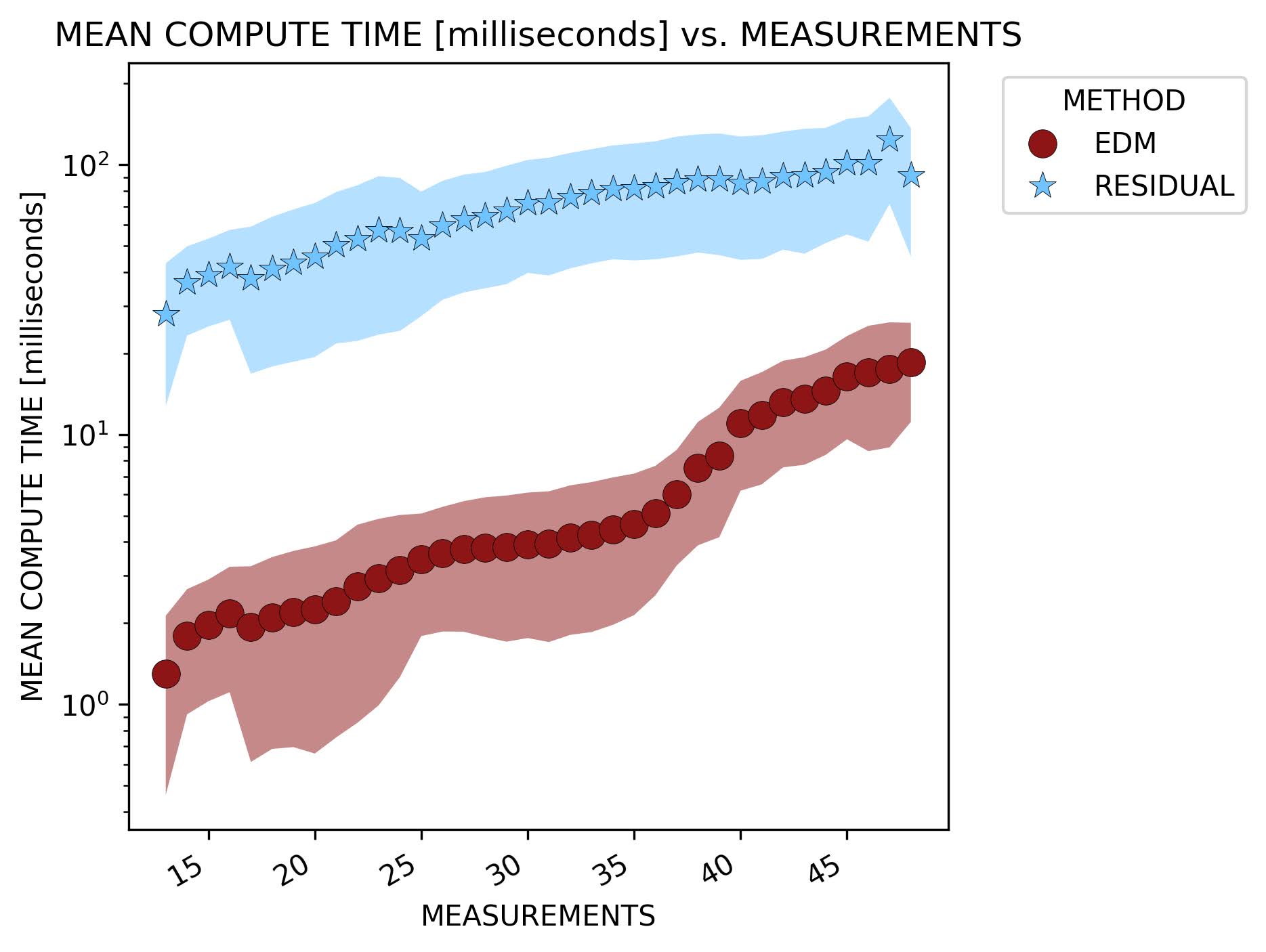}
    \captionof{figure}{Mean computation time of greedy EDM and greedy residual FDE with increasing number of measurements. One standard deviation on either side of the mean is shaded.}
    \label{fig:time_measurements}
\end{figure}

\end{minipage}%
\hfill
\begin{minipage}[b]{0.46\textwidth}

\begin{figure}[H]
    \raggedleft
    \centering
    \includegraphics[width=\linewidth]{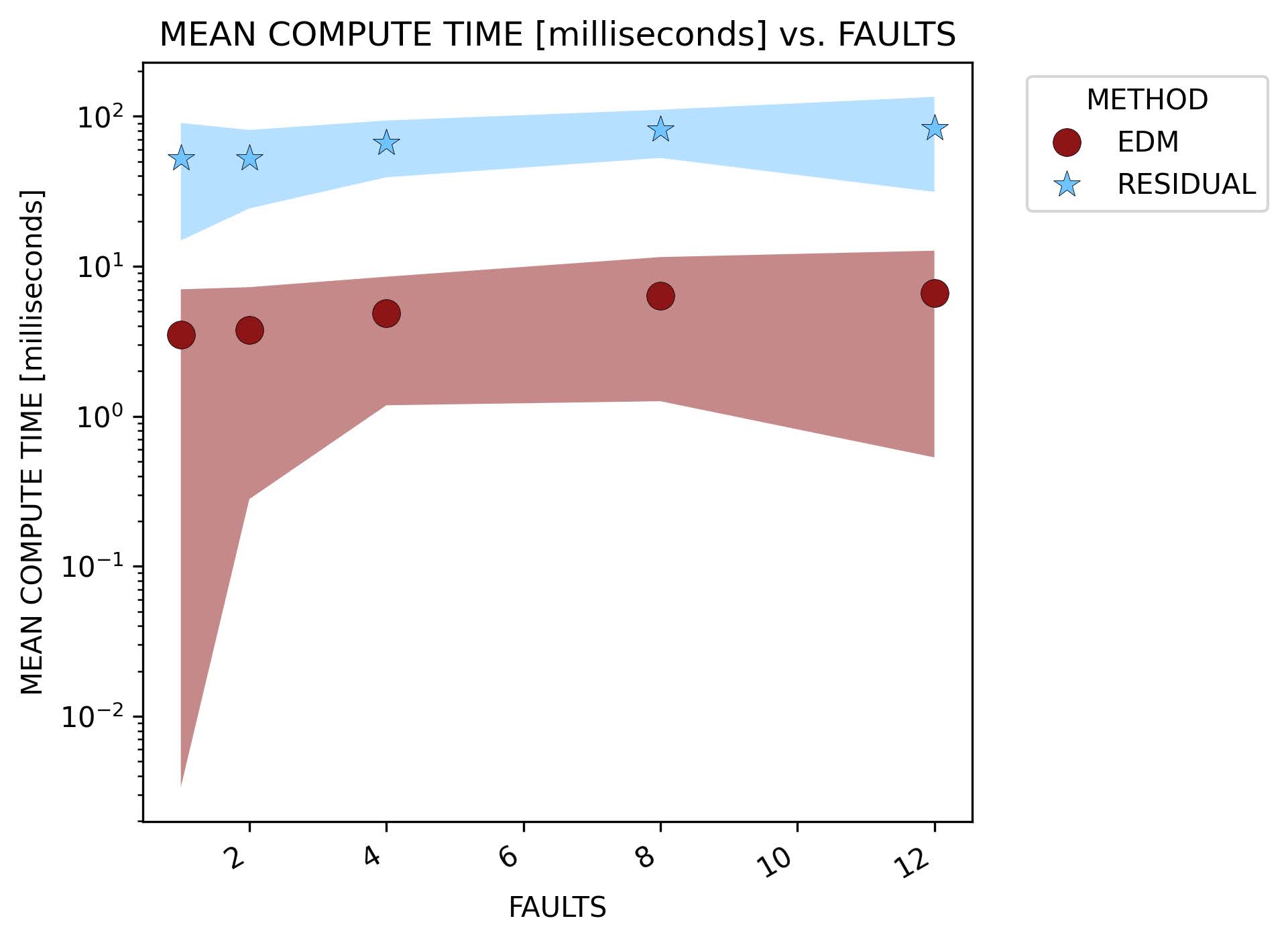}
    \captionof{figure}{Mean computation time of greedy EDM and greedy residual FDE with increasing number of faults. One standard deviation on either side of the mean is shaded.}
    \label{fig:time_faults}
\end{figure}

\end{minipage}%
\end{flushleft}

\subsection{Fault Exclusion Accuracy}
\label{sec:simulated_accuracy}
\begin{figure}[H]
    \centering
    \includegraphics[height=8cm]{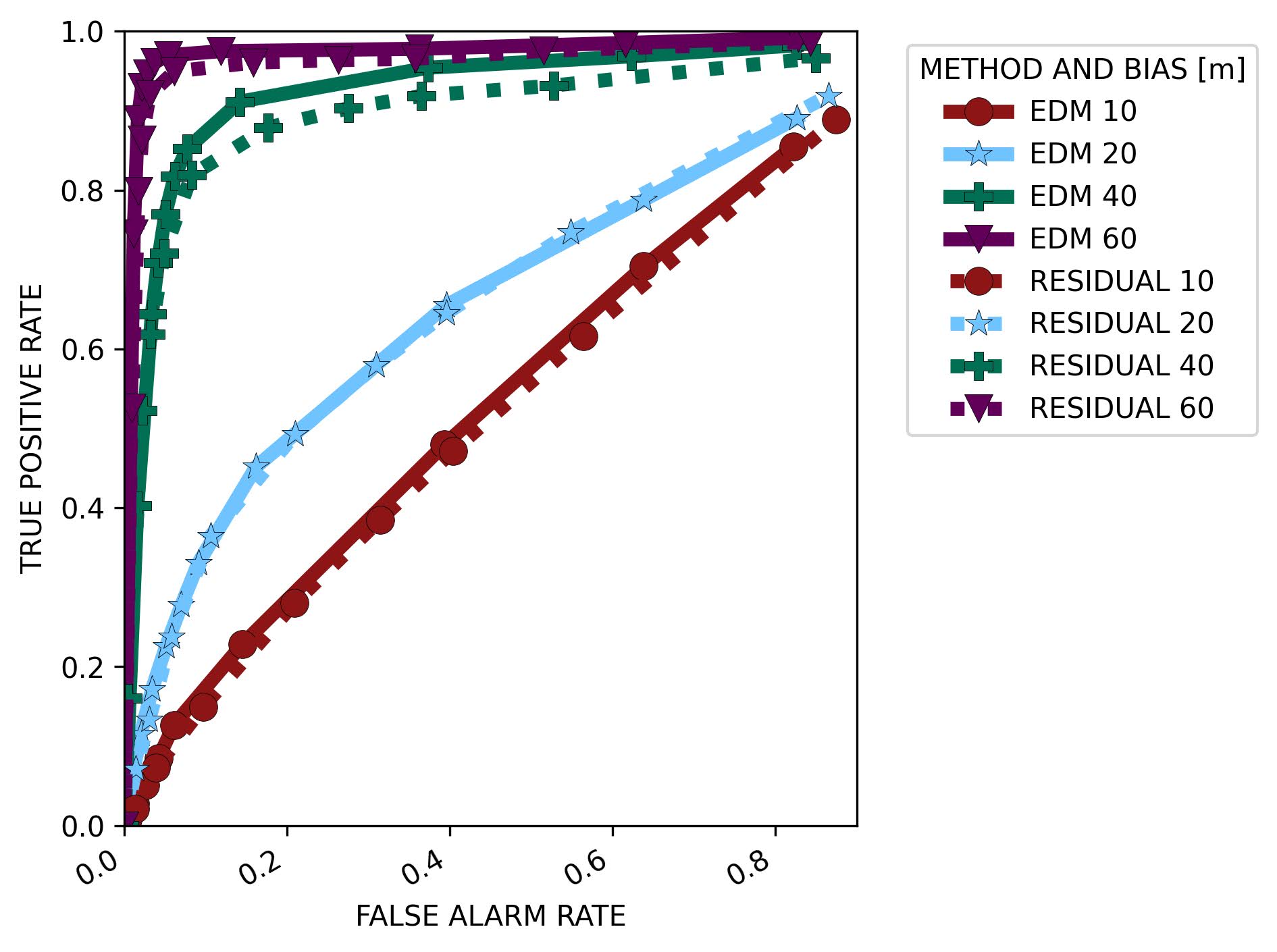}
    \captionof{figure}{The receiver operating characteristic (ROC) curve for both greedy EDM FDE and greedy residual FDE across four different fault bias magnitudes.}
    \label{fig:roc_curve}
\end{figure}

Our measure for fault exclusion accuracy uses the receiver operating characteristic (ROC) curve. Figure \ref{fig:roc_curve} shows the ROC curves for both greedy EDM and greedy residual FDE for a scenario with eight faults at the Munich location and varying fault magnitudes. The ROC curve is created by sweeping across a range of threshold parameters and shows how the true positive rate (the percentage of faulty measurements correctly identified) increases as the false alarm rate (percentage of non-faulty measurements incorrectly identified as faulty) increases. An algorithm is deemed to be more accurate at excluding faults if the curve goes towards the top left corner of the graph. Figure \ref{fig:roc_curve} shows that both algorithms are better at detecting faults as the magnitude of the fault increases  above the noise floor of the 10m zero mean Gaussian noise added to all measurements. Figure \ref{fig:roc_curve} also shows that greedy EDM and greedy residual FDE perform nearly identically for each fault bias magnitude.
A single quantitative metric from the ROC curve is calculated as the area under the curve (AUC) for each scenario. A larger AUC means the method is better at excluding faults.

In Table \ref{table:auc}, we show the area under the curve for all locations and two conditions of a single 20m fault bias and eight 60m fault biases. The area under the curve (AUC) is comparable for both greedy residual and greedy EDM FDE illustrating the fact that greedy EDM FDE is able to achieve comparable performance in terms of fault detection accuracy across the dataset when compared to greedy residual FDE. As described in Section \ref{sec:simulated} and displayed in Figure \ref{fig:in_view}, the Calgary, London, and Zurich locations were created using a larger elevation mask leaving fewer visible satellites. Since those locations have fewer total measurements, the AUC at those locations is lower when eight measurements are faulty because there are fewer measurements available to create a consistent fault-free measurement set.

\begin{table}[h]
\caption{Area under the receiver operating characteristic (ROC) curve for both greedy EDM and greedy residual FDE.}
\label{table:auc}
\centering
\begin{tabular}{rcccc}
\multicolumn{1}{l}{}                  &                             &                                                & \multicolumn{2}{c}{\cellcolor[HTML]{EFEFEF}\textbf{Area Under ROC Curve}} \\ \cline{4-5} 
\rowcolor[HTML]{EFEFEF} 
\textbf{Location}                     & \textbf{Fault Bias {[}m{]}} & \textbf{\# Faults}                             & \textbf{Greedy EDM}               & \textbf{Greedy Residual}              \\ \hline
\rowcolor[HTML]{EFEFEF} 
Calgary                               & \cellcolor[HTML]{C0C0C0}20  & \multicolumn{1}{c|}{\cellcolor[HTML]{EFEFEF}1} & \textbf{0.44}                     & \textbf{0.44}                        \\ \hline
\rowcolor[HTML]{EFEFEF} 
\cellcolor[HTML]{C0C0C0}Cape Town     & \cellcolor[HTML]{C0C0C0}20  & \multicolumn{1}{c|}{\cellcolor[HTML]{EFEFEF}1} & \textbf{0.51}                    & 0.48                                 \\ \hline
\rowcolor[HTML]{EFEFEF} 
Hong Kong                             & \cellcolor[HTML]{C0C0C0}20  & \multicolumn{1}{c|}{\cellcolor[HTML]{EFEFEF}1} & \textbf{0.50}                    & 0.48                                 \\ \hline
\rowcolor[HTML]{EFEFEF} 
\cellcolor[HTML]{C0C0C0}London        & \cellcolor[HTML]{C0C0C0}20  & \multicolumn{1}{c|}{\cellcolor[HTML]{EFEFEF}1} & 0.42                             & \textbf{0.43}                        \\ \hline
\rowcolor[HTML]{EFEFEF} 
Munich                                & \cellcolor[HTML]{C0C0C0}20  & \multicolumn{1}{c|}{\cellcolor[HTML]{EFEFEF}1} & \textbf{0.49}                    & 0.48                                 \\ \hline
\rowcolor[HTML]{EFEFEF} 
\cellcolor[HTML]{C0C0C0}Sao Paulo     & \cellcolor[HTML]{C0C0C0}20  & \multicolumn{1}{c|}{\cellcolor[HTML]{EFEFEF}1} & \textbf{0.52}                    & 0.51                                 \\ \hline
\rowcolor[HTML]{EFEFEF} 
Stanford Oval                         & \cellcolor[HTML]{C0C0C0}20  & \multicolumn{1}{c|}{\cellcolor[HTML]{EFEFEF}1} & \textbf{0.51}                    & 0.49                                 \\ \hline
\rowcolor[HTML]{EFEFEF} 
\cellcolor[HTML]{C0C0C0}Sydney        & \cellcolor[HTML]{C0C0C0}20  & \multicolumn{1}{c|}{\cellcolor[HTML]{EFEFEF}1} & 0.51                             & \textbf{0.52}                        \\ \hline
\rowcolor[HTML]{EFEFEF} 
Zurich                                & \cellcolor[HTML]{C0C0C0}20  & \multicolumn{1}{c|}{\cellcolor[HTML]{EFEFEF}1} & 0.46                             & \textbf{0.47}                        \\ \hline
\rowcolor[HTML]{EFEFEF} 
\cellcolor[HTML]{C0C0C0}Calgary       & 60                          & \multicolumn{1}{c|}{\cellcolor[HTML]{C0C0C0}8} & 0.26                             & \textbf{0.30}                        \\ \hline
\rowcolor[HTML]{EFEFEF} 
Cape Town                             & 60                          & \multicolumn{1}{c|}{\cellcolor[HTML]{C0C0C0}8} & \textbf{0.67}                    & 0.65                                 \\ \hline
\rowcolor[HTML]{EFEFEF} 
\cellcolor[HTML]{C0C0C0}Hong Kong     & 60                          & \multicolumn{1}{c|}{\cellcolor[HTML]{C0C0C0}8} & 0.69                             & \textbf{0.70}                        \\ \hline
\rowcolor[HTML]{EFEFEF} 
London                                & 60                          & \multicolumn{1}{c|}{\cellcolor[HTML]{C0C0C0}8} & 0.28                             & \textbf{0.33}                        \\ \hline
\rowcolor[HTML]{EFEFEF} 
\cellcolor[HTML]{C0C0C0}Munich        & 60                          & \multicolumn{1}{c|}{\cellcolor[HTML]{C0C0C0}8} & \textbf{0.68}                    & 0.67                                 \\ \hline
\rowcolor[HTML]{EFEFEF} 
Sao Paulo                             & 60                          & \multicolumn{1}{c|}{\cellcolor[HTML]{C0C0C0}8} & \textbf{0.67}                    & 0.66                                 \\ \hline
\rowcolor[HTML]{EFEFEF} 
\cellcolor[HTML]{C0C0C0}Stanford Oval & 60                          & \multicolumn{1}{c|}{\cellcolor[HTML]{C0C0C0}8} & \textbf{0.67}                    & 0.66                                 \\ \hline
\rowcolor[HTML]{EFEFEF} 
Sydney                                & 60                          & \multicolumn{1}{c|}{\cellcolor[HTML]{C0C0C0}8} & \textbf{0.69}                    & \textbf{0.69}                                 \\ \hline
\rowcolor[HTML]{EFEFEF} 
\cellcolor[HTML]{C0C0C0}Zurich        & 60                          & \multicolumn{1}{c|}{\cellcolor[HTML]{C0C0C0}8} & 0.27                             & \textbf{0.32}                        \\ \hline
\end{tabular}
\end{table}

\section{Real World Data Results}
\label{sec:gsdc}

To further validate the proposed greedy EDM FDE algorithm, we also compare our algorithm against greedy residual FDE and our previous 2021 version of EDM FDE on the Google Smartphone Decimeter Challenge (GSDC) 2023 \citep{smartphone-decimeter-2023}. We use a subset of the total dataset by taking the 68 urban traces from seven different android phones that include multipath indicators. This dataset consistently provides 25-45 satellite measurements at each timestep epoch. In Section~\ref{sec:gsdc_time} we present results of the compute time on the real-world data and in Section~\ref{sec:gsdc_accuracy} we provide results on the localization accuracy achievable using either our new greedy EDM FDE method, residual FDE, or our previous 2021 EDM FDE algorithm.

Since EDM FDE is reliant on matching distances between pairs of points, EDM FDE works best if those distances are as close to their truth value as possible. Unlike when using the simulated data in Section~\ref{sec:simulated}, when using real-world data, we first condition the pseudorange by removing known biases from the measured pseudorange using the equation: 
\begin{align}
    \label{eq:model}
    \rho = \rho_m + b_{rx} - I + T - c_b,
\end{align}
Where $\rho_m$ is the measured pseudorange, $b_{rx}$ is the receiver's clock bias, $I$ is the Ionosphere delay, $T$ is the Troposphere delay, and $c_b$ is the constellation bias. In the results that follow, we estimated the receiver clock bias using weighted least squares and the atmospheric and constellation bias are provided in the GSDC dataset itself.

\subsection{Computation Time}
\label{sec:gsdc_time}
Greedy EDM FDE outperforms greedy residual FDE in terms of computation speed by an order of magnitude when averaged across all 68 real-world data traces. Figure~\ref{fig:gsdc_time} shows the average computation time for all FDE methods with increasing number of measurements at each timestep. One standard deviation on either side of the mean is shaded on the plot. This plot justifies the fact that Greedy EDM FDE can rapidly remove multiple faults in the presence of many measurements. Both our new EDM FDE method presented in this paper and our previous 2021 version of EDM FDE perform similarly in terms of computation speed.

Since both EDM FDE and residual FDE algorithms greedily remove faults, the computation time of both algorithms is not only a factor of the number of measurements but also the number of faults predicted. If the threshold for either algorithm is set low, then the algorithm will more often enter the fault exclusion step and iterate again to remove more satellite measurements. If the threshold for either algorithm is set high, then the algorithm will rarely enter the fault exclusion step nor iterate to remove additional measurements. To remove the effect the threshold value has on computation time, for each trace, we used the computation time of the threshold that minimized the accuracy metric we describe in Section~\ref{sec:gsdc_accuracy}.

\begin{flushleft}     
\begin{minipage}[b]{0.46\textwidth}

\begin{figure}[H]
    \centering
    \includegraphics[height=5.8cm]{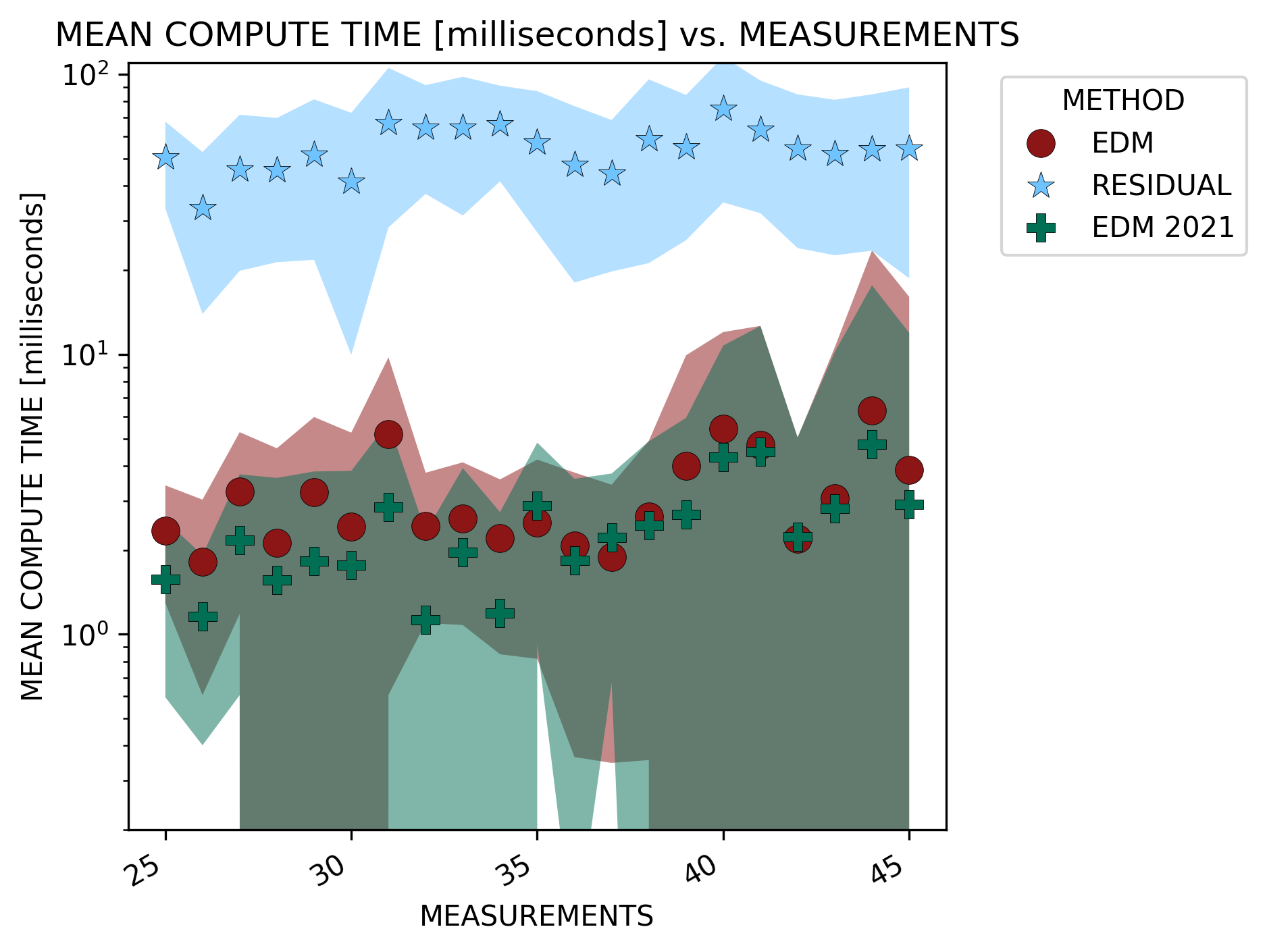}
    \captionof{figure}{Mean computation time of greedy EDM and greedy residual FDE with increasing number of measurements. One standard deviation on either side of the mean is shaded.}
    \label{fig:gsdc_time}
\end{figure}

\end{minipage}%
\hfill
\begin{minipage}[b]{0.46\textwidth}

\begin{figure}[H]
    \centering
    \includegraphics[height=6.2cm]{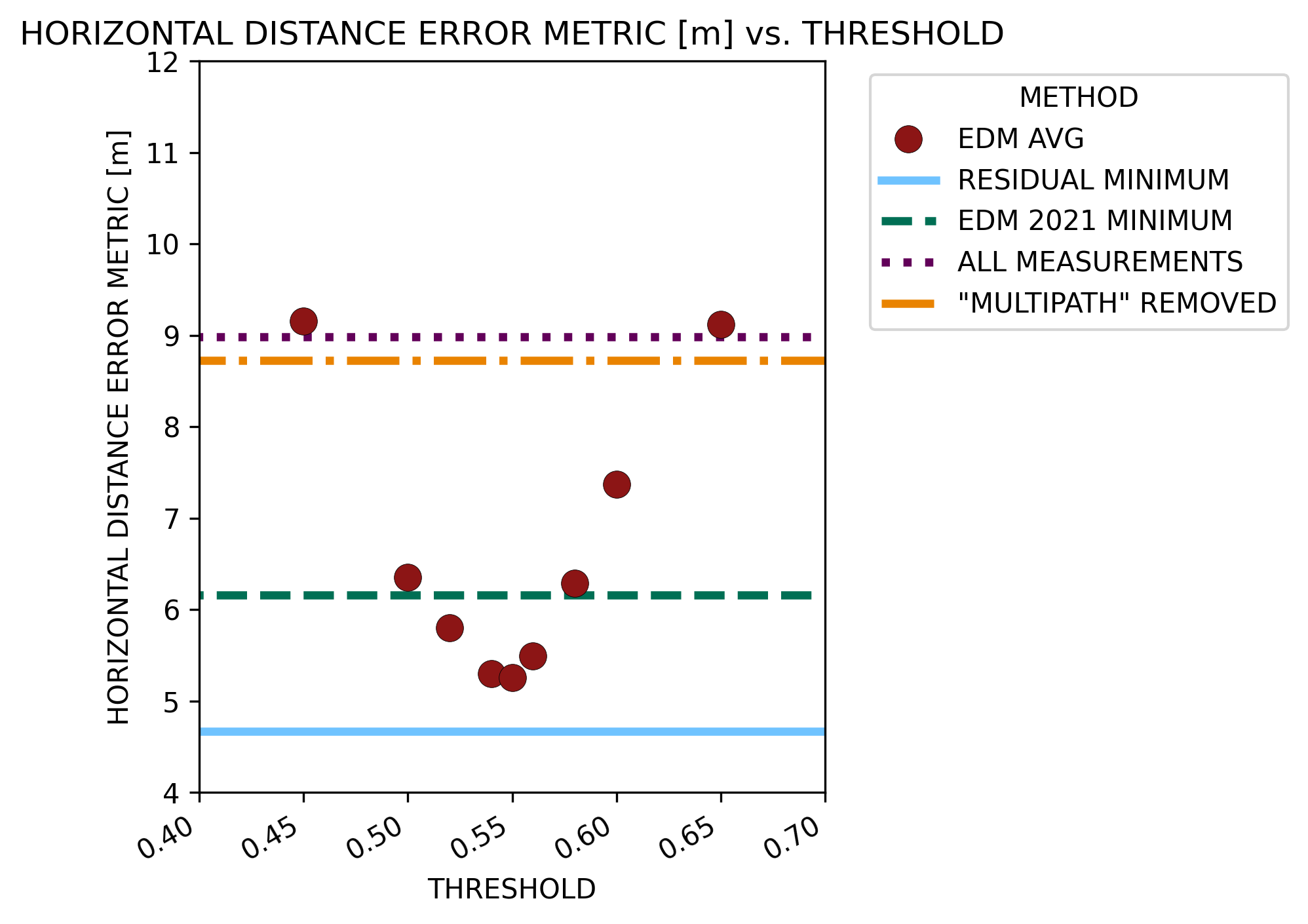}
    \captionof{figure}{Horizontal distance error metric for multiple choices of the EDM FDE threshold.}
    \label{fig:gsdc_edm}
\end{figure}

\end{minipage}%
\end{flushleft}

\subsection{Localization Accuracy}
\label{sec:gsdc_accuracy}
In the real world, there is no clear consensus of what constitutes a ``faulty" satellite measurement. The GSDC dataset includes indicators of whether the android operating system deemed the satellite measurement to be multipath, but provides no information on whether measurements came from line-of-sight satellites, etc. To avoid rigidly defining ``faults" as multipath, non-line-of-sight, or something else, we abstract away satellite level faults and instead focus on the localization accuracy using the predicted non-faulty measurements.

The GSDC 2023 dataset provides a ``ground truth" position at each timestep \citep{smartphone-decimeter-2023}. For each algorithm, we predict which satellite measurements are faulty, then run weighted least squares to estimate the receiver's position with the non-faulty measurements. We then use the same horizontal distance error metric as the GSDC \citep{smartphone-decimeter-2023} by computing the 50th and 95th percentile horizontal distance error of the weighted least squares solution when compared with the GSDC ``ground truth" for each trace. The 50th and 95th percentiles are then averaged together to reduce to a single value for each trace. We perform this process for each of the 68 real-world traces and then average this horizontal distance error metric across all the traces. We repeat this method sweeping across a range of threshold values for each algorithm.

Figure~\ref{fig:gsdc_edm} shows the horizontal distance error metric for our new EDM FDE algorithm across a range of threshold values from 0.45 to 0.65. The plot shows that using EDM FDE achieves the lowest errors when the threshold value is set to the middle of that range near 0.55. If the threshold is too high, then missed detections result in larger errors. If the threshold is set too low, then too many satellite measurements are removed thus degrading the geometry of the localization problem and increasing error. The plot also compares against weighted least squares using all measurements and using all measurements which the android operating system has deemed are not multipath measurements. EDM FDE is thus shown to decrease horizontal position errors by correctly removing faulty satellite measurements. Figure~\ref{fig:gsdc_edm} also plots the minimum error achievable with the other two FDE algorithms of greedy residual FDE and our previous 2021 version of EDM FDE. The plot shows that our new version of EDM FDE from this paper is able to outperform our previous 2021 version of EDM FDE. In Section~\ref{sec:edm_improvements} we described our change to the fault detection test statistic that provides a more clear indication of measurement faults despite noise in the measurements. Thus, our new version of EDM FDE from this paper is able to more accurately remove faults in the noisy GSDC dataset and achieve a lower localization error than our previous 2021 version of EDM FDE.

Figure~\ref{fig:gsdc_residual} shows the horizontal distance error metric when using residual-based FDE for threshold values from 10 to 100,000 and Figure~\ref{fig:gsdc_edm_2021} shows the horizontal distance error metric when using our previous 2021 version of EDM FDE for threshold values from 0.3 to 300. Greedy residual FDE is able to achieve the lowest horizontal position errors with a well-chosen threshold and our previous 2021 version of EDM FDE has the largest horizontal distance errors out of the three algorithms we tested even with the best threshold values.
Figures~\ref{fig:gsdc_edm}--~\ref{fig:gsdc_edm_2021} do show that no matter the FDE method, lower horizontal distance errors are possible by removing faults versus using all measurements or only removing the measurements that the android operating system indicated include multipath.
\begin{flushleft}     
\begin{minipage}[b]{0.46\textwidth}

\begin{figure}[H]
    \centering
    \includegraphics[width=\linewidth]{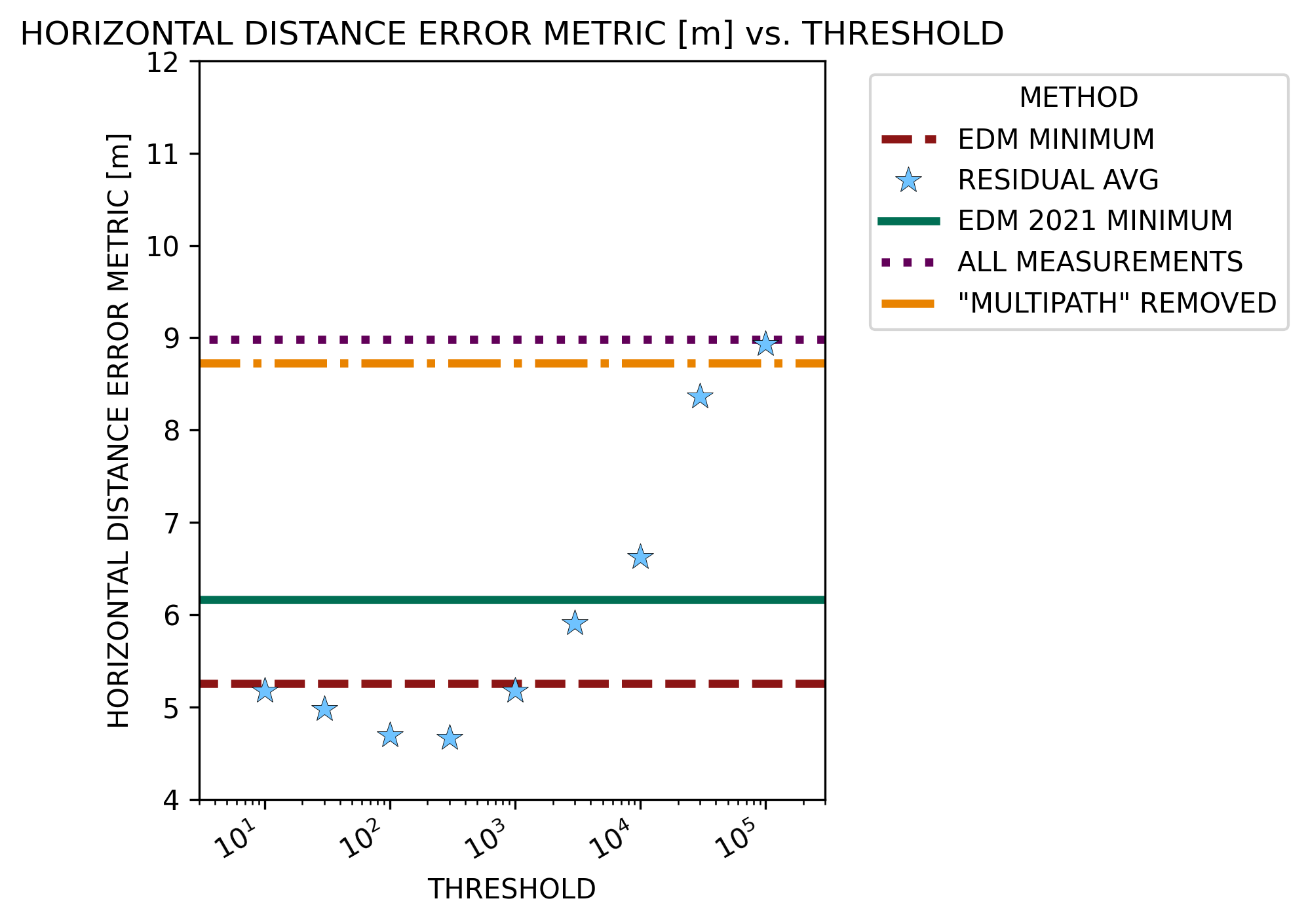}
    \captionof{figure}{Horizontal distance error metric for multiple choices of the residual FDE threshold.}
    \label{fig:gsdc_residual}
\end{figure}

\end{minipage}%
\hfill
\begin{minipage}[b]{0.46\textwidth}

\begin{figure}[H]
    \raggedleft
    \centering
    \includegraphics[width=\linewidth]{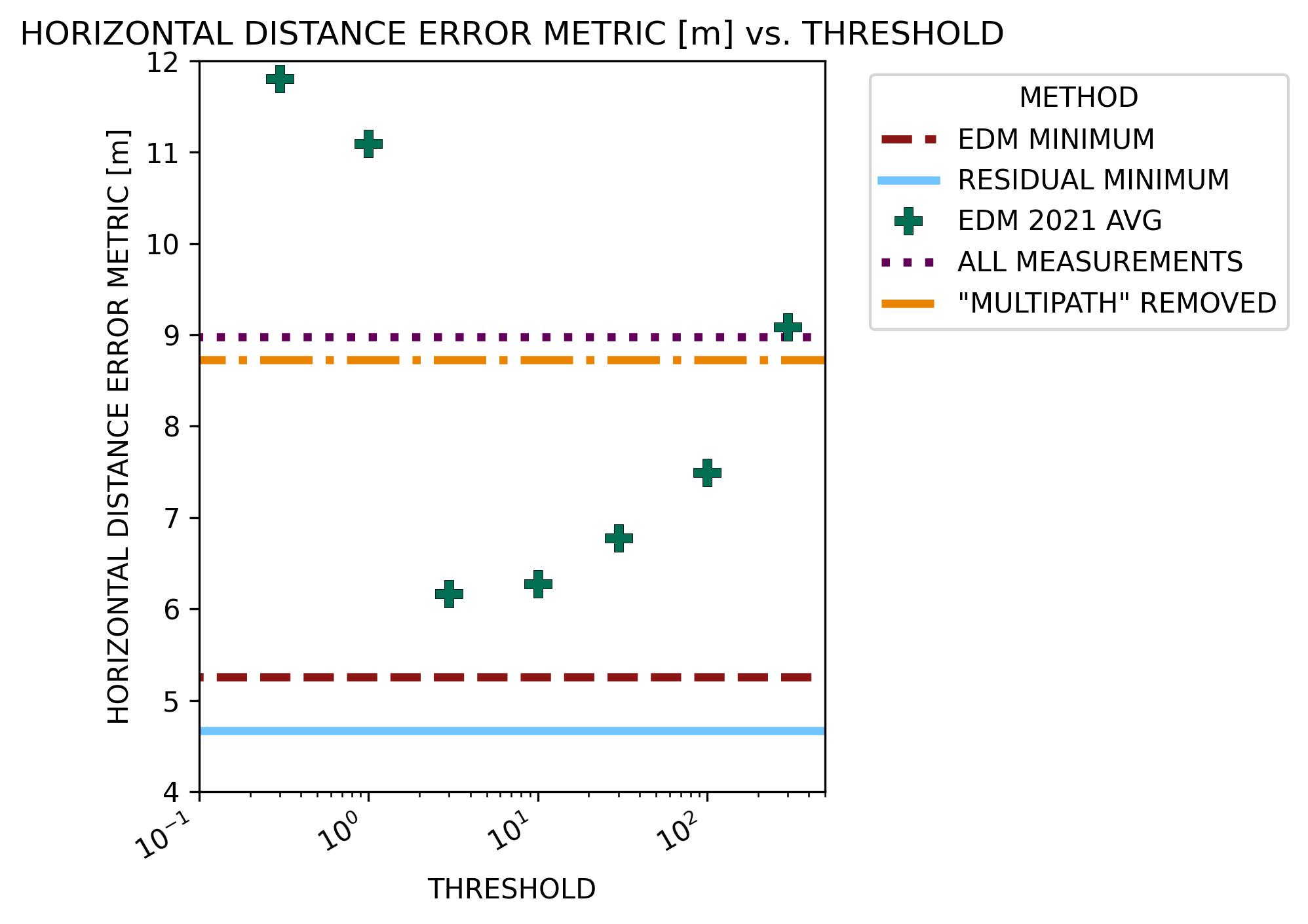}
    \captionof{figure}{Horizontal distance error metric for multiple choices of the EDM 2021 threshold.}
    \label{fig:gsdc_edm_2021}

\end{figure}
\end{minipage}%
\end{flushleft}

\section{Time Complexity Derivations}
\label{app:time}

This section presents a theoretical runtime complexity comparison for EDM FDE in Section~\ref{sec:edm_alg_time}, residual FDE in Section~\ref{sec:residual_alg_time} and solution separation in Section~\ref{sec:solution_separation_alg_time}. We offer general remarks and a visual comparison to finalize in Section~\ref{app:final_compare}. For ease of comparison, we use common notation found in Section~\ref{app:notation} for all algorithms.

\subsection{Nomenclature}
\label{app:notation}
\nomenclature[01]{\(m\)}{Number of satellite pseudorange measurements (integer)}
\nomenclature[02]{\(f\)}{Number of predicted faults (integer)}
\nomenclature[03]{\(\rho\)}{raw pseudorange measurements corrected for atmospheric effects, satellite clock bias, and constellation/signal clock bias (column vector of size $m \times 1$)}
\nomenclature[04]{\(X_{sv}\)}{positions (ECEF) of satellites (matrix of size $m \times 3$)}
\nomenclature[05]{\(B_{sv}\)}{clock bias of satellites (column vector of size $m \times 1$)}
\nomenclature[06]{\(X_{r}\)}{position (ECEF) of receiver (matrix of size $1 \times 3$)}
\nomenclature[07]{\(X_{r_i}\)}{position (ECEF) of receiver with the $i$th measurement removed (matrix of size $1 \times 3$)}
\nomenclature[08]{\(W\)}{Weights for each measurement as a square, diagonal matrix (matrix of size $m \times m$)}
\nomenclature[99]{\(T\)}{detection threshold (float)}
\printnomenclature

\subsection{EDM FDE}
\label{sec:edm_alg_time}

The EDM FDE algorithm analyzed in this section is shown in Algorithm~\ref{alg:edm-fde}. The most compute heavy step of the algorithm is the eigenvalue decomposition step which takes $\mathcal{O}\left((m+1)^3\right)$ time at the first step since the matrix has size of the number of measurements plus one for the receiver. The size of the matrix for eigenvalue decomposition is one unit smaller each time a measurement fault is removed. The total runtime complexity to remove all $f$ faults can thus be summarized as
\begin{equation}
    \mathcal{O}\left( \sum_{a=0}^{f} (m+1-a)^3 \right)
\end{equation}

\subsection{Greedy Residual FDE}
\label{sec:residual_alg_time}

 Most residual-based algorithms have greatly diverged from the original residual FDE algorithm proposed in the 1980s \citep{PARKINSON1988}. Instead, most residual-based algorithms are a greedy version that iteratively remove the measurement with the largest normalized residual. The residual FDE baseline approach implemented in this paper is a version of greedy~residual~FDE~\citep{blanch2015fast}.

 The main idea of residual-based FDE is that the weighted least square receiver position ($X_r$) is calculated with all measurements and then using that position estimate the pseudorange residuals ($R$) and chi-square test statistic ($\chi^2$) is computed. If the chi-square test statistic is above the provided detection threshold then the pseudorange measurement with the largest normalized pseudorange residual ($\tilde{R}$) is removed until either there are only four satellites measurements remaining or the chi-square test statistic falls below the provided detection threshold.
The greedy residual algorithm is outlined in Algorithm~\ref{alg:greedy_residual}, Algorithm~\ref{alg:chi_square}, and Algorithm~\ref{alg:norm_residual}.

The longest runtime operation for residual FDE is the matrix inversion that happens during weighted least squares to solve for the position estimate used in the residuals calculation and again during the chi square test statistic calculation. The matrix inversion for the chi square test statistic only happens once for each fault removed, but the matrix inversion for weighted least squares may happen up to a maximum of 20 times based on how long it takes the Newton-Raphson implementation to converge. Matrix inversion takes $\mathcal{O}(m^3)$ for the first fault and then the size of the matrix inverse is one unit smaller each time a measurement fault is removed. In total the runtime complexity of removing all $f$ faults is thus:
\begin{equation}
    \mathcal{O}\left( \sum_{a=0}^{f} k(m-a)^3 \right),
\end{equation}
where $k$ may take on a value between two and 21 based on how many iterations Newton-Raphson takes to converge.

\begin{flushleft}     
\begin{minipage}[t]{0.45\textwidth}
	\begin{algorithm}[H]
    	\caption{Greedy Residual FDE}
    	\label{alg:greedy_residual}
    	\begin{algorithmic}[1]
    	    
            \Inputs{}
                \Indent
                \State $\rho$ : raw pseudorange measurements corrected for atmospheric effects, satellite clock bias, and constellation/signal clock bias
                \State $W$ : Weights for each measurement
                \State $X_{sv}$ : positions (ECEF) of satellites
                \State $B_{sv}$ : clock bias of satellites
                \State $T$ : detection threshold
                \EndIndent
            \Initialize{}
                \Indent
                \State $X_r \leftarrow \text{weighted\_least\_squares}(\rho,X_{sv},B_{sv},W)$
                \State $R \leftarrow \rho - \|X_r - X_{sv} \|$
                \State $\tau \leftarrow \text{test\_statistic}(X_r,X_{sv},R,W)$
                \EndIndent
                \While{$\tau > T$ and \# satellites $>$ 4}
                    \State $i$ $\leftarrow$ normalized\_residual($X_r,X_{sv},R,W$)
                    \State delete $i$ from $\rho$, $X_{sv}$, $B_{sv}$, and $W$
                     \State $X_r \leftarrow \text{weighted\_least\_squares}(\rho,X_{sv},B_{sv},W)$
                    \State $R \leftarrow \rho - \|X_r - X_{sv} \|$
                    \State $\tau \leftarrow \text{test\_statistic}(X_r,X_{sv},R,W)$
                \EndWhile
    	\end{algorithmic}
    \end{algorithm}
\end{minipage}%
\hfill
\begin{minipage}[t]{0.45\textwidth}
	\begin{algorithm}[H]
    	\caption{test\_statistic: Chi-Square Test Statistic}
    	\label{alg:chi_square}
    	\begin{algorithmic}[1]
    	    
            \Inputs{}
                \Indent
                \State $X_r$ : Receiver's position in ECEF x, y, and z
                \State $X_{sv}$ : Satellites' positions in ECEF x, y, and z
                \State $R$ : pseudorange residuals for each measurement
                \State $W$ : Weights for each measurement
                \EndIndent
            \Initialize{}
                \Indent
                \State $G \leftarrow \frac{X_r - X_{sv}}{\|X_r - X_{sv}\|}$
                \EndIndent
            \State $\chi^2 \leftarrow R^\top \left(W - WG(G^\top W G)^{\inv})G^\top W\right) R$
            \State \Return $\chi^2$
    	\end{algorithmic}
    \end{algorithm}

	\begin{algorithm}[H]
    	\caption{normalized\_residual: Largest Normalized Residual}
    	\label{alg:norm_residual}
    	\begin{algorithmic}[1]
    	    
            \Inputs{}
                \Indent
                \State $X_r$ : Receiver's position in ECEF x, y, and z
                \State $X_{sv}$ : Satellites' positions in ECEF x, y, and z
                \State $R$ : pseudorange residuals for each measurement
                \State $W$ : Weights for each residual
                \EndIndent
            \Initialize{}
                \Indent
                \State $ n \leftarrow$ number of satellite measurements
                \State $G \leftarrow \frac{X_r - X_{sv}}{\|X_r - X_{sv}\|}$
                \State $\tilde{x} \leftarrow (G^\top W G)^{\inv}G^\top W R$
                \EndIndent
                
            \For{index $i$ in range($n$)}
                \State $\tilde{r}_i \leftarrow \frac{w_i(r_i-g_i^\top\tilde{x})^2}{1 - g_i^\top w_i(G^\top W G)^{\inv}g_i}$
            \EndFor
            
            \State index $\leftarrow \arg \max(\tilde{r}_i \forall \; i$ in range($m$))
            
            \State \Return index
    	\end{algorithmic}
    \end{algorithm}

\end{minipage}%
\end{flushleft}

\subsection{Solution Separation FDE}
\label{sec:solution_separation_alg_time}

The solution separation algorithm analyzed in this section is a synthesis of previously proposed solution separation algorithms from the literature \citep{brown1988, Joerger2014, Pullen2021, joerger2016, blanch2015}. The solution separation algorithm is outlined in Algorithm~\ref{alg:solution_separation}, Algorithm~\ref{alg:ss_norm}, and Algorithm~\ref{alg:ss_thresholds}.

At its heart, solution separation computes a position estimate for all possible subset combinations for the detection step and then reiterates that same test in the exclusion step to verify that outliers were properly removed. The number of subset combinations to be tested is a combination factor summing the combination of removing one fault out of the $m$ measurement choices up to the combinations of removing $f$ faults out of the $m$ measurement choices. The number of total subset combinations is thus:
\begin{equation}
    {\sum_{a=1}^{f}} {_m}C_a = {\sum_{a=1}^{f}} \frac{m!}{a!(m-a)!}.
\end{equation}
The most compute heavy step of computing a position solution using weighted least squares is the matrix inversion step that takes $\mathcal{O}(m^3)$ time. Thus putting together the two steps of detection and exclusion, the total runtime can be approximated by:
\begin{equation}
    \mathcal{O}\left( m^3 \left( {1 + \sum_{a=1}^{f}} \frac{m!}{a!(m-a)!} \right)  + (m-1)^3 \left( {1 + \sum_{a=1}^{f-1}} \frac{(m-1)!}{a!(m-a-1)!} \right)\right).
\end{equation}

\subsection{Theoretical Runtime Comparison Summary}
\label{app:final_compare}
The true computation time of each algorithm depends on its specific implementation including programming language, CPU and GPU quality, method chosen for matrix inversion and eigenvalue decomposition, data structures, parallelization, etc. The step that dominates the asymptotic time complexity for greedy residual FDE is the matrix pseudoinversion that takes place while using weighted least squares to solve for the receiver's position estimate after each fault is excluded. The step that dominates the asymptotic time complexity for greedy EDM FDE is the eigenvalue decomposition of the Gram matrix. While both computing a pseudoinverse and eigenvalue decomposition have an asymptotic time complexity of $\mathcal{O}(m^3)$, in our Python implementation using \texttt{NumPy} \citep{2020NumPy-Array}, the eigenvalue decomposition of the Gram matrix is significantly faster. Eigenvalue decomposition with \texttt{NumPy} uses the Linear Algebra Package (LAPACK) routine called \_gesdd which is able to rapidly perform the eigenvalue decomposition using a divide and conquer strategy \citep{netlib} thanks in part to the fact that the Gram matrix is positive semidefinite with a symmetric structure. Additionally, residual FDE may need to calculate the pseudoinverse multiple times for Newton Raphson to converge versus EDM FDE which does not have to iteratively compute the eigenvalue decomposition multiple times.

To provide a visual comparison of the asymptotic runtime complexity of the three algorithms, we plug in several possible measurement values and fault values into the runtime complexity equations that were derived in Sections~\ref{sec:edm_alg_time}~--~\ref{sec:solution_separation_alg_time}. For the runtime of residual FDE, we choose a value of $k=10$ which is in the middle of its possible range between two and 21. Figure~\ref{fig:faults_4} shows the runtime complexity value computed by varying the number of measurements for four faults. Figure~\ref{fig:faults_16} shows the runtime complexity value computed by varying the number of measurements for 16 faults. The figures show that solution separation takes several orders of magnitude longer in terms of computation time when comparing the theoretical asymptotic runtime complexity. Residual FDE is also theoretically two to 20 times slower than EDM FDE depending on the value of $k$.

\begin{flushleft}     
\begin{minipage}[b]{0.46\textwidth}

\begin{figure}[H]
    \centering
    \includegraphics[width=\linewidth]{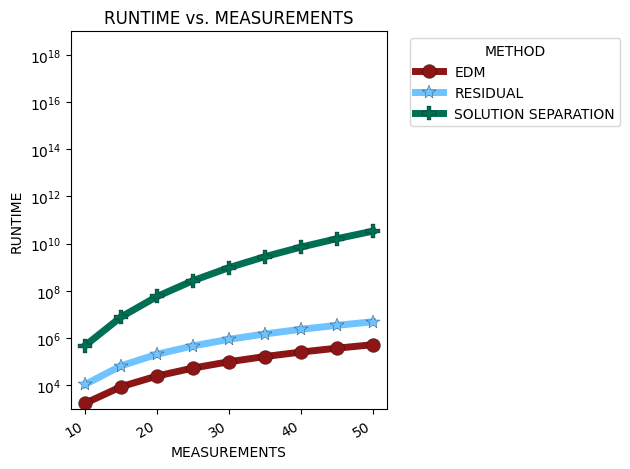}
    \captionof{figure}{Runtime comparison with four faults.}
    \label{fig:faults_4}
\end{figure}

\end{minipage}%
\hfill
\begin{minipage}[b]{0.46\textwidth}

\begin{figure}[H]
    \raggedleft
    \centering
    \includegraphics[width=\linewidth]{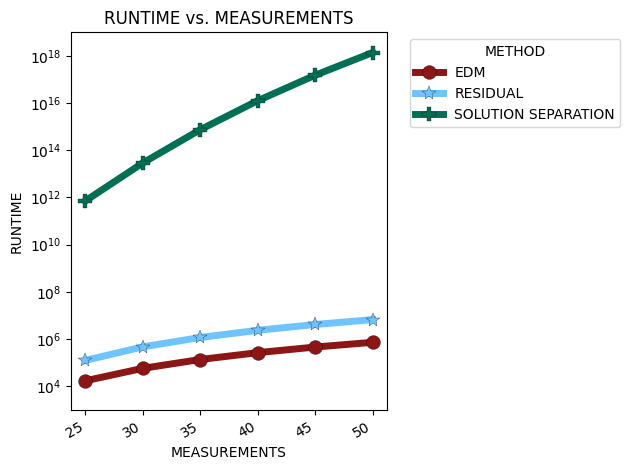}
    \captionof{figure}{Runtime comparison with 16 faults.}
    \label{fig:faults_16}
\end{figure}
\end{minipage}%
\end{flushleft}

\begin{flushleft}     
\begin{minipage}[t]{0.45\textwidth}
	\begin{algorithm}[H]
    	\caption{Solution Separation FDE}
    	\label{alg:solution_separation}
    	\begin{algorithmic}[1]
    	    
            \Inputs{}
                \Indent
                \State $\rho$ : raw pseudorange measurements corrected for atmospheric effects, satellite clock bias, and constellation/signal clock bias
                \State $W$ : Weights for each measurement
                \State $X_{sv}$ : positions (ECEF) of satellites
                \State $B_{sv}$ : clock bias of satellites
                \State $T_{params}$ : detection threshold parameters $\beta$, $P_{Hi}$, $C_{REQ}$ from \cite{joerger2016}
                \EndIndent
            \Initialize{}
                \Indent
                \State $ n \leftarrow$ number of satellite measurements
                \State $X_r \leftarrow \text{weighted\_least\_squares}(\rho,X_{sv},B_{sv},W)$
                \State $\sigma^2 \leftarrow \text{ss\_normalizer}(X_{sv},X_r)$
                \EndIndent
            \If{\# satellites $<$ 6}
                \State \Return none
            \EndIf
                
            \For{indexes $i$ in range(all combinations)}
                \State delete $i$ from $\rho$, $X_{sv}$, $B_{sv}$, and $W$
                
                \State $X_{r_i} \leftarrow \text{weighted\_least\_squares}(\rho_{m},X_{sv},B_{sv},W)$
                \State $\sigma_i^2 \leftarrow \text{ss\_normalizer}(X_{sv},X_r)$
                \State $\Delta_i \leftarrow \|X_r - X_{r_i} \|$
                \State $\sigma_{\Delta i} \leftarrow \sqrt{\sigma_i^2 - \sigma^2}$
                \State $q_i \leftarrow \Delta_i / \sigma_{\Delta i}$
                \State $T_d$, $T_e \leftarrow \text{ss\_thresholds}(T_{params})$
                \If{$q_i > T_d$}
                    \State Repeat SS FDE recursively using $\rho$, $X_{sv}$, $B_{sv}$, and $W$ and $T_e$ as threshold.
                    \If{$q_i < T_e$}
                        \State fault index $\leftarrow i$
                    \EndIf
                \EndIf
            \EndFor

            \State \Return {fault index}
    	\end{algorithmic}
    \end{algorithm}
\end{minipage}%
\hfill
\begin{minipage}[t]{0.45\textwidth}
	\begin{algorithm}[H]
    	\caption{ss\_normalizer}
    	\label{alg:ss_norm}
    	\begin{algorithmic}[1]
    	    
            \Inputs{}
                \Indent
                \State $X_r$ : Receiver's position in ECEF x, y, and z
                \State $X_{sv}$ : Satellites' positions in ECEF x, y, and z
                \EndIndent
            \State $G \leftarrow \frac{X_r - X_{sv}}{\|X_r - X_{sv}\|}$
            \State $\sigma^2 \leftarrow trace\left((G^\top G)^{\inv}\right)$
            \State \Return $\sigma^2$
    	\end{algorithmic}
    \end{algorithm}

	\begin{algorithm}[H]
    	\caption{ss\_thresholds}
    	\label{alg:ss_thresholds}
    	\begin{algorithmic}[1]
    	    
            \Inputs{}
                \Indent
                \State $T_{params}$ : detection threshold parameters $\beta$, $P_{Hi}$, $C_{REQ}$ from \cite{joerger2016}
                \EndIndent
            \State $T_d \leftarrow$ compute using Equation (41) from \cite{joerger2016}
            \State $T_e \leftarrow$ compute using Equation (42) from \cite{joerger2016}
            \State \Return $T_d$, $T_e$
    	\end{algorithmic}
    \end{algorithm}

\end{minipage}%
\end{flushleft}

\section{Modifications to Greedy EDM FDE}
\label{sec:mods}
Similar to greedy residual FDE, there are several common modifications that can be performed for greedy EDM FDE to decrease its runtime or add theoretical guarantees. None of these discussed modifications were used in the presented results, but could reduce the computation time of either greedy algorithm.

The first potential modification is that one could remove multiple faults at the same time. Removing multiple of the satellite measurements with the largest absolute mean in the $n+1$ and $n+2$ eigenvectors as discussed in Section \ref{sec:exclusion} means that fewer greedy fault exclusion iterations need to be performed and can thus decrease total runtime at each timestep.

The second potential modification is that if a maximum number of faults is known, either greedy FDE algorithm could be stopped preemptively if the maximum number of faults has been reached. This modification may prevent non-faulty measurements from incorrectly being excluded.

The third potential modification is that the detection threshold could be set automatically if the measurement data can be post processed. As described in Section \ref{sec:detection}, adding noise to the measurements increases the pair of the $n+1$ and $n+2$ eigenvalues. This means that the amount of noise in the GNSS measurements should change the test statistic threshold at which a fault is detected. The distribution of the fault detection test statistic could be calculated across all measurement timesteps in the dataset. The new detection threshold could then be set as the lowest value for the detection test statistic. This process could then repeated until the distribution of the test statistic has converged across all measurement timesteps.

Another potential modification is more tightly coupling the EDM FDE method with a temporal localization filter. In order for EDM FDE to perform well, the pseudorange measurements should be as close as possible to the true distance between the receiver the the satellite positions. In the GSDC dataset, we added an additional preprocessing step to subtract out the android phone's receiver clock bias. Modifying the algorithm to pull the receiver clock bias, atmospheric effect variables, etc. from a temporal filter running in parallel might also increase the efficiency of either algorithm.

Finally, if probabilistic guarantees (e.g. protection levels) are desired, one could run solution separation after the EDM FDE algorithm is complete \citep{Pullen2021}. This method would have the benefit of removing measurement outliers rapidly using EDM FDE while also providing protection levels using the well-characterized solution separation formal method.

\section{Conclusion}
\label{sec:conclusion}

This paper presents a greedy fault detection and exclusion algorithm using Euclidean distance matrices.
The fault detection test statistic and fault exclusion strategy is described in detail. 
We present results on both a simulated and real-world dataset showing that greedy EDM FDE outperforms greedy residual FDE in terms of computation time while still maintaining high exclusion accuracy and low horizontal position errors.

\section*{Acknowledgments}
The authors thank the members of the Stanford NAV Lab for discussions that contributed to this work and especially feedback from Dr. Tara Mina, Dr. Shubh Gupta, Keidai Iiyama, and Marta Cortinovis. This work was supported by the National Science Foundation (NSF) under award number 2006162 and by NASA Jet Propulsion Laboratory’s (JPL’s) Strategic University Research Partnership program.

\printbibliography[title=References]

\end{document}